\documentclass{article}

\usepackage{arxiv}
\usepackage[T1]{fontenc}    % use 8-bit T1 fonts
\usepackage{booktabs}       % professional-quality tables
\usepackage{makecell}
\usepackage{amsmath}
\usepackage{amsfonts}       % blackboard math symbols
\usepackage{graphicx}
\usepackage{doi}
\usepackage{subfig}
\usepackage{xcolor}
\usepackage{hyperref}       % hyperlinks
\usepackage{url}            % simple URL typesetting
\usepackage{setspace}

\date{}
\renewcommand{\headeright}{}%{Technical Report}
\renewcommand{\undertitle}{}%{Technical Report}
\renewcommand{\shorttitle}{}%{\textit{arXiv} Template}

\hypersetup{
pdftitle={CALText: Contextual Attention Localization for Offline Handwritten Text},
pdfsubject={cs.CV, cs.LG, cs.NE},
pdfauthor={Tayaba Anjum, Nazar Khan},
pdfkeywords={Handwriting, Urdu, Offline, Recognition, Text, Attention, Localization, Encoder, CNN, DenseNet, Decoder , Recurrent, GRU, LSTM, OCR},
}

\title{CALText: Contextual Attention Localization for Offline Handwritten Text}

\author{Tayaba Anjum \\
	Department of Computer Science\\
	University of the Punjab\\
	Lahore, Pakistan \\
	\texttt{tayaba.anjum@pucit.edu.pk} \\
	\And
	\href{https://orcid.org/0000-0002-9470-2120}{\includegraphics[scale=0.06]{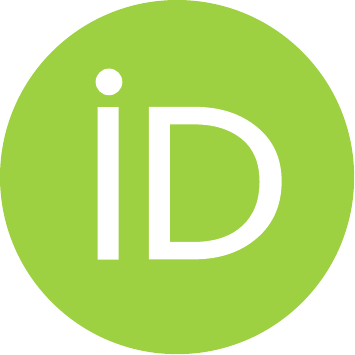}\hspace{1mm}Nazar Khan}
	\thanks{\url{http://faculty.pucit.edu.pk/nazarkhan/}}
	\\
	Department of Computer Science\\
	University of the Punjab\\
	Lahore, Pakistan \\
	\texttt{nazarkhan@pucit.edu.pk} \\
}

\begin{document}
\maketitle

\begin{abstract}
Recognition of Arabic-like scripts such as Persian and Urdu is more challenging than Latin-based scripts. This is due to the presence of a two-dimensional structure, context-dependent character shapes, spaces and overlaps, and placement of diacritics. Not much research exists for offline handwritten Urdu script which is the 10th most spoken language in the world. We present an attention based encoder-decoder model that learns to read Urdu in context. A novel localization penalty is introduced to encourage the model to attend only one location at a time when recognizing the next character. In addition, we comprehensively refine the only complete and publicly available handwritten Urdu dataset in terms of ground-truth annotations. We evaluate the model on both Urdu and Arabic datasets and show that contextual attention localization outperforms both simple attention and multi-directional LSTM models.
\end{abstract}


%\linenumbers
\doublespacing

%% main text
\section{Introduction}
\label{sec:introduction}
Despite being the 10th most spoken language in the world \cite{ethnologue200}, there has been very little focus on the offline recognition of handwritten Urdu text. As a spoken language, Urdu is similar to Hindi but in script form, it is closer to Arabic and Persian. Offline Handwritten Text Recognition (OHTR) refers to the problem of recognizing handwritten text from an image. OHTR provides the interface between humans and machines by automatic digitization of handwritten documents such as letters, lecture notes, bank checks, medical records, and library books. The OHTR problem varies from Optical Character Recognition (OCR) because of the wide variety of writing styles, sizes, sentence lengths, pen types, page backgrounds, and rule violations.

\begin{figure}[!t]
    \centering
    \includegraphics[width=.75\linewidth]{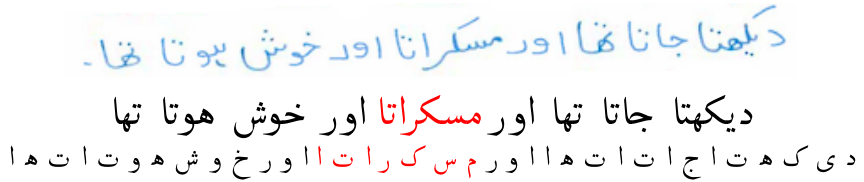}
    \caption{\textbf{Top}: Handwritten Urdu text. \textbf{Middle}: Corresponding typed version. \textbf{Bottom}: Isolated typed characters appear significantly different from their corresponding cursive ligatures. Urdu is neither written nor typed in the form of isolated characters.}
    \label{fig:texttype_comparison}
\end{figure}

For Latin-based handwritten text, good quality datasets are readily available \cite{Marti2002, LeCun1998a, Taghva1999, Mouchere2013}. Similar datasets exist for handwritten Arabic \cite{Abed2002, Alamri2016, Amara2005, Tong2013, Mahmoud2014, ahmed2019handwritten} and Persian \cite{sadri2016novel} as well. Recognition of Arabic-like scripts (Arabic, Persian and Urdu) presents interesting challenges in addition to those for Latin-based scripts \cite{Naz2015}. For example, the presence of two-dimensional structure, character and ligature overlap, ligature deformation, context-sensitive shapes of ligatures, placement and quantities of diacritical marks, stretching, and spacing. Figure \ref{fig:texttype_comparison} depicts handwritten Urdu text in the first row, the corresponding typed text in the second row and the individual characters in the third row. It can be seen that within cursively written Urdu words, the individual characters appear with widely varying ligatures. While typed Urdu conforms to some rules, handwriting can be restricted to very few rules.

Previous attempts at OHTR for Arabic-like scripts have either used raw pixels or features extracted from a Convolutional Neural Network (CNN) that are then decoded using a recurrent model such as a bidirectional Long Short-Term Memory (BLSTM) network \cite{ahmed2019handwritten, Hassan2019} that aims to capture the horizontal nature of \emph{deskewed} text. Multidimensional LSTM (MDLSTM) has also been employed to capture the horizontal as well as vertical nature of Arabic-like scripts \cite{Gui2018, Ahmed2017}. These approaches use the complete image for recognition of a character at a certain time step. However, a single character can be recognized from a specific location of an image at a certain time step using attention. Attention mechanism helps to recognize a character from an image by focusing on a specific region at a certain time step.

In previous efforts for Arabic-like scripts, error reporting has been restricted to the level of characters. World level accuracy is not reported. This, we speculate, is due to the character level accuracy not being good enough for recognition of words since a word is recognized correctly if and only if each and every character is recognized correctly.

Traditional variants of LSTM (1 direction), BLSTM (2 directions) and MDLSTM (4 directions) all limit the number of directions that a decoder can proceed in. In contrast, an attention-based decoder is direction-less since it can focus on any relevant part of the image at any given time. However, characters in a sequence can’t appear at arbitrary locations. For text, \emph{attention itself needs context}. That is, where to focus next depends on where the model has already focused previously. One such decoder was introduced in \cite{anjum_icfhr2020}. In solving problems related to recognition of a line of text, each character appears within a small contiguous region instead of being scattered all over the image. Therefore, in addition to the context of attention,  text requires localization of attention as well. We show in this work how localization can be introduced into a contextual attention-based decoder. This results in our proposed \emph{contextual attention localization} (CAL) method for handwritten text.

In this paper we present a thorough explanation of the encoder/decoder architecture from \cite{anjum_icfhr2020}, extend it using attention localization and introduce a technique for compact spatiotemporal attention visualization. We use a DenseNet encoder to extract high level features from the input image. The features are then decoded into a contextual sequence of output characters using localized attention. Using localized attention in context encourages our model to attend only to relevant image regions while decoding an output sequence. The encoder and decoder are trained jointly so that the decoder can guide the encoder towards generation of richer, more useful features. The paper contributes as follows:
\begin{enumerate}
\item We present a detailed contextual attention-based, encoder-decoder model for the recognition of offline handwritten text. In particular, we describe the model introduced in \cite{anjum_icfhr2020} in much greater detail and extend it.
\item We show that, armed with attention, direction-less decoders out-perform uni-directional, bidirectional and multi-directional decoders.
\item We show that for text recognition, encouraging the model to localize its attention improves recognition accuracy.
\item We comprehensively re-annotate the PUCIT-OHUL dataset \cite{anjum_icfhr2020} of offline handwritten Urdu text using 7 annotators.
\begin{itemize}
    \item Missing ground-truth lines are added.
    \item We synchronize ground-truth text with the text on the images.
    \item We record instances of text that is crossed-out, overwritten, or covered with correction fluid.
    \item We annotate incorrectly spelled words as they are. We do not correct spelling mistakes in the written text.
\end{itemize}
\item We additionally report word level accuracy instead of only character level accuracy.
\item To show the robustness of the proposed model, we use Urdu PUCIT-OHUL and Arabic KHATT dataset for experiments.
\item We introduce a method for compact visualization of spatiotemporal attention.
\item Compared to previous approaches, we report $2\times$ accuracy improvement at both character level and word level.
\end{enumerate}
For reproducible research, code for this work is publicly available at \url{https://github.com/nazar-khan/CALText}.

The rest of the paper is structured as follows. Section \ref{sec:literature_review} reviews the state-of-the-art related to offline handwritten text recognition. Challenges specific to Urdu and other Arabic-like scripts are discussed in Section \ref{sec:challenges}. This is followed by detailed explanations of our encoding and decoding frameworks in Section \ref{sec:methodology}. Section \ref{sec:datasets} describes the datasets used for training and evaluation. Quantitative as well as qualitative experimental results are presented in Section \ref{sec:exp_results} and the paper is concluded in Section \ref{sec:conclusion}.

\section{Literature Review}
\label{sec:literature_review}
Deep convolutional networks with some form of recurrence, attention or self-attention represent state-of-the-art for the OHTR problem for multiple scripts such as Latin-based \cite{coquenet_icfhr2020_gated_fcn, ly_icdar2021_self_attention_crn}, Arabic \cite{Ahmed2017, Hassan2019}, CJK\footnote{Chinese Japanese Korean} \cite{xiao_icdar2019_chinese, ly_icfhr2020_japanese, nguyen_icfhr2020_japanese} and even mathematical expressions \cite{li_icfhr2020_math} and music notes \cite{baro_icfhr2020_music}. Historically, OHTR has evolved from convolutional \cite{LeCun1998a} to recurrent \cite{Graves2009} to attention-based \cite{Gui2018} models. %\cite{Graves2009, Ahmed2017, Hassan2019}.
The use of attention remains a central theme in other problems involving text, such as scene text detection \cite{wu_jvcir_2021_attention_rcnn_text, wu_jvcir2021_attention_alignment_text}.

Early efforts at Urdu recognition were restricted to typed documents with fixed font styles and sizes \cite{Javed2009b, Akram2014}. Uni-, bi- and multi-directional LSTMs \cite{Hasan2013, Hasan2015, Naz2016, Naz2017a, Naz2017b} were also applied for typed Urdu recognition.

Handwritten Urdu text recognition is more difficult compared to printed text as writing style, pen type, page background, and rule violations heavily impact the recognition process. Early attempts were naturally restricted to recognition of isolated characters. Isolated Urdu numerals were recognized in \cite{Naz2010} using a Support Vector Machine (SVM) classifier with Quad-Tree based Longest-Run (QTLR) features. The CENPARMI-U dataset introduced in \cite{sagheer2009cenparmi_urdu} focused on isolated characters and contained 44 handwritten Urdu characters and numerals, 5 special characters, and only 57 Urdu words. It also contained Urdu dates and numeral strings. Isolated characters were recognized in \cite{Pathan2012} using an SVM classifier of moment invariant features. For the recognition of isolated words in online Urdu handwritten text, a Hidden Markov Model (HMM) with fuzzy logic was used in \cite{Razzak2010} to achieve $74.1\%$ accuracy on the IFN/ENIT dataset for Urdu Naskh style and $87.6\%$ accuracy for Urdu Nastaleeq style. In \cite{Mukhtar2010}, isolated words were recognized using gradient and structural features with SVM classifier. They created a dataset of 100 words written by 2 writers. Each word was written 8 times by 1 writer to obtain a total of 1600 images.

Deep learning based solutions for offline handwritten Urdu text include \cite{Ahmed2019} and \cite{Hassan2019} which both employ bi-directional LSTM models. The only publicly available dataset is UCOM/UNHD \cite{Ahmed2019} and that too is only partially available.

\section{Challenges of OHTR}
\label{sec:challenges}
\begin{figure}[!t]
    \centering
    \includegraphics[width=.5\linewidth]{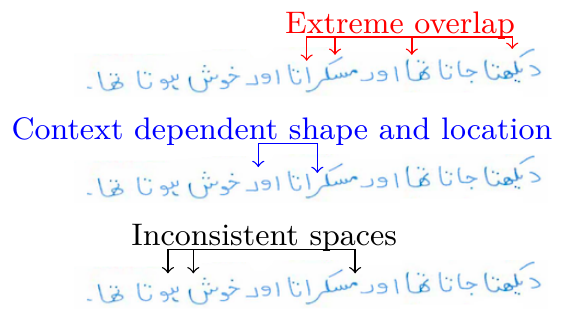}
\caption{Challenges in offline handwritten Arabic-like scripts. This Urdu sentence contains i) extreme overlap between ligatures due to the 2-dimensional structure of Urdu, ii) context dependent ligature shapes as well as locations, and iii) intra-word spaces occasionally larger than inter-word spaces.}
\label{fig:urdu_challenges}
\end{figure}

Recognition of Arabic-like scripts such as Urdu is difficult due to Sayre's paradox \cite{sayre1973machine} that a cursively written word cannot be recognized without being segmented and cannot be segmented without being recognized. In addition, handwritten text can employ a wide variety of pen types, writing styles, sizes and page backgrounds. But most importantly, in handwritten Urdu and text in general, almost every rule can be violated until the text becomes illegible. Some of the main challenges in offline handwritten Urdu text recognition as explained in Figure \ref{fig:urdu_challenges} are:
\begin{enumerate}
\item One challenge is that cursive Urdu can have extreme overlap among characters. Sometimes, parts of a letter can overlap multiple other letters or even other words. In other cases, letters can lie exactly on top of other letters.
\item Another challenge is context-dependent shape and location of letters. The placement of characters is not just in right-to-left order. Urdu script has a 2-dimensional structure as well.
\item A third challenge is that Urdu is not entirely cursive. It is a mix of cursive and non-cursive. As a result, sometimes the space within characters of a word can be larger than the space between words. This makes tokenization of a sentence into words difficult.
\item Within cursive words, characters to be recognized often appear with widely varying ligatures. While typed Urdu conforms to some rules, handwriting can be restricted to very few rules.
\end{enumerate}
Therefore, handwritten Urdu and Arabic recognition is not a trivial problem.

\section{Methodology}
\label{sec:methodology}
\begin{figure}[!t]
    \centering
    \includegraphics[width=.5\linewidth]{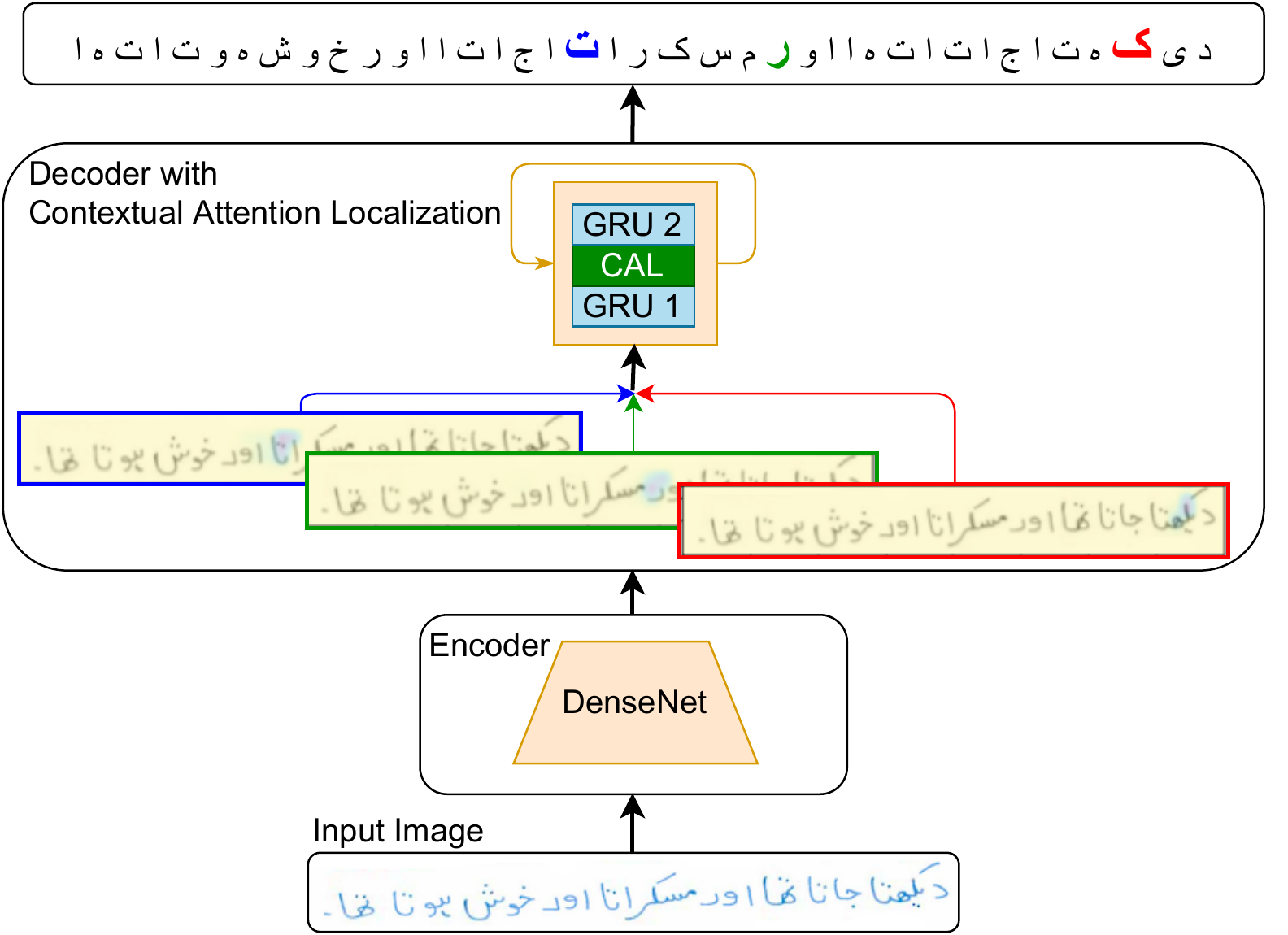}
    \caption{Proposed encoder/decoder framework. An image containing handwritten text is input to a DenseNet encoder. Extracted features are passed through a decoder that attends to specific image regions in context so as to produce probabilities of output characters at every time step. The probabilities are finally decoded into output characters using beam search.}
    \label{fig:architecture}
\end{figure}

For handwritten text, even if the input images are restricted to have a fixed size, the characters may constitute a variable-length sequence.  The encoder/decoder model with a recurrent decoder can be used for the recognition of such variable text sequences \cite{Cho2014}. The encoder transforms an input image into an intermediate representation. The decoder then transforms the intermediate representation into a sequence of output characters. While decoding, an attention mechanism is used to focus on a specific part of the image to produce different characters. An overview of our approach is shown in Figure \ref{fig:architecture}. Given an input image, the model produces a variable-length sequence of 1-hot vectors
\begin{equation}
    Y = \{\mathbf{y}_1 , \mathbf{y}_2, \dots,  \mathbf{y}_T\}
    \label{eq:1hot_sequence}
\end{equation}
where $\mathbf{y}_t  \in \mathbb{R}^k$ is the $1$-of-$k$ encoding of the $t$-th character in the image text containing $T$ characters including any blank spaces and punctuation marks. Size of the vocabulary is denoted by $k$.

\subsection{Encoder: DenseNet with Bottlenecks}
\begin{figure}[!b]
\centering
\includegraphics[width=.5\linewidth]{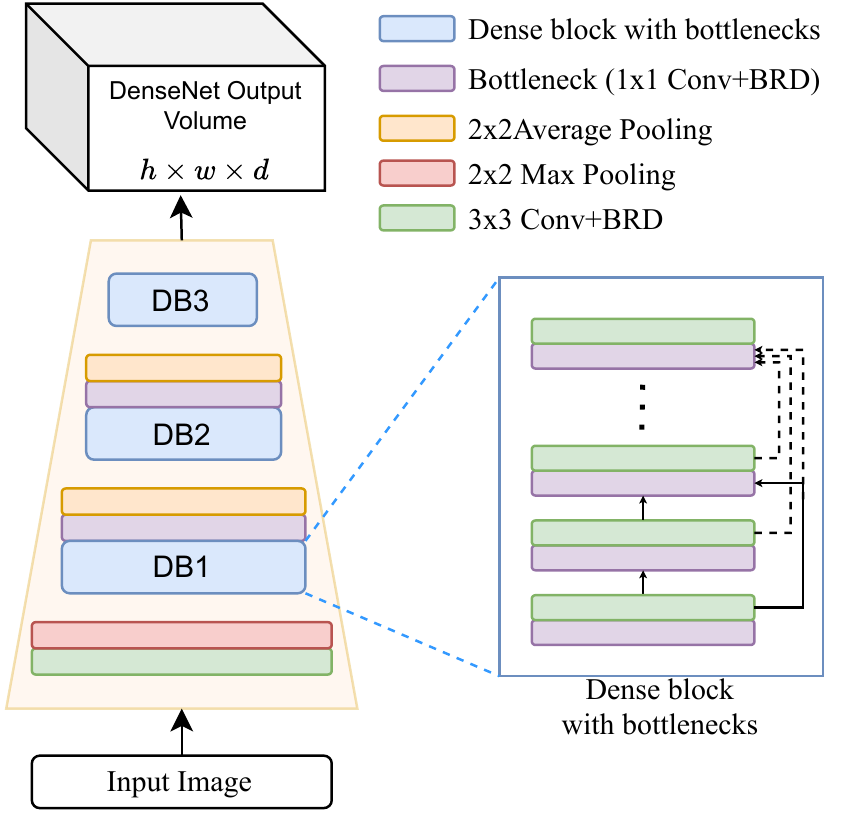}
\caption{Architecture of the DenseNet encoder that transforms an input image into a 3-dimensional feature volume. BRD = Batchnorm + ReLU + Dropout.}
\label{fig:DFCN}
\end{figure}

For extraction of high-level salient features, we use the DenseNet \cite{huang2017_densenet} architecture which is a variation of CNNs with dense blocks. In a dense block, convolution is applied to all previous layers within the block instead of just the previous layer. The use of dense blocks provides a more diverse set of features by minimizing information loss during forward and backward propagation. Since connecting all previous layers becomes costly, bottleneck layers ($1\times1$ convolutions) are inserted to compress the volume of previous layers. The convolutional output volume of layer $i$ is calculated as
\begin{equation}
    {\mathbf{F}_i} = {\mathcal{C}_i}(\mathcal{B}_i([\mathbf{F}_1;\mathbf{F}_2;\dots; \mathbf{F}_{i-1}]))
    \label{eq:conv_output_volume}
\end{equation} 
where $\mathcal{C}_i(\cdot)$ represents regular convolution layers, $\mathcal{B}_i(\cdot)$  represents $1 \times 1$ convolution of a bottleneck layer  and $[\mathbf{F}_1;\mathbf{F}_2;\dots; \mathbf{F}_{i-1}]$ represents the depth-wise concatenation of the outputs of all previous layers. Pooling layers between dense blocks are used for extraction of features at multiple scales. The detailed architecture of our DenseNet encoder is shown in Figure \ref{fig:DFCN}. There are $3$ dense blocks. Each block consists of $32$ sub-layers alternating between $1 \times 1$ convolution bottleneck layers and $3 \times 3$ convolution layers. Each dense block is preceded by a $2\times 2$ pooling layer with stride $2$.

\begin{figure}[!t]
    \centering
    \includegraphics[width=.6\linewidth]{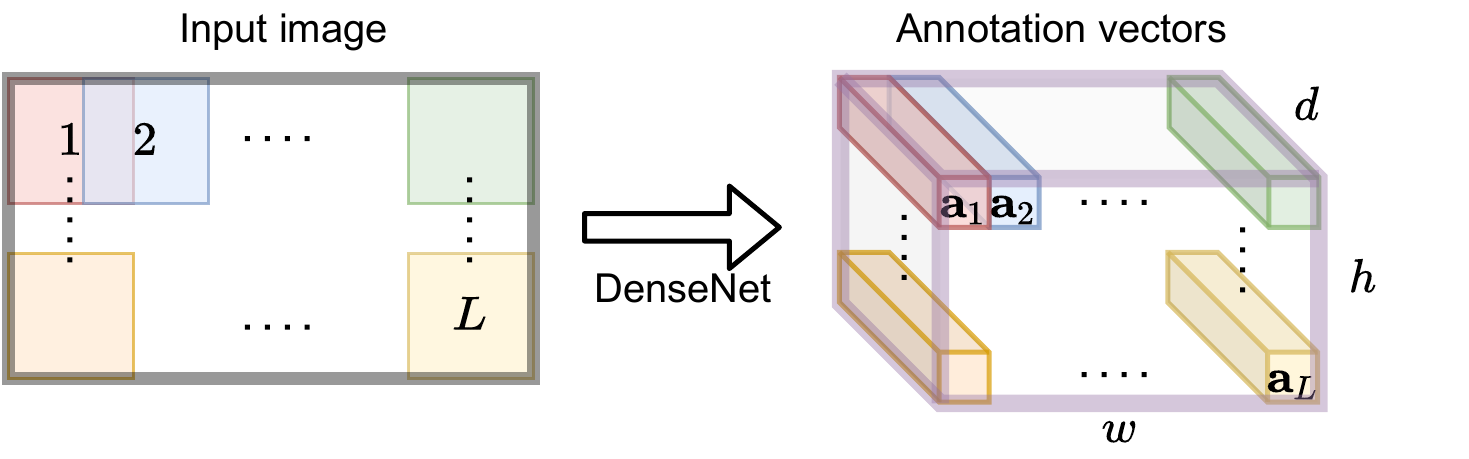}
    \caption{DenseNet output volume of size $h \times w \times d$ represents $d$-dimensional \emph{annotation vectors} of $h \times w$ overlapping blocks of the input image.}
\label{fig:annotation_vectors}
\end{figure}

The output of the dense encoder, for a given input image, is a volume of dimensions $h \times w \times d$. It can be interpreted as $d$-dimensional \emph{annotation vectors} of $h \times w$ overlapping blocks of the input image. This interpretation is illustrated in Figure \ref{fig:annotation_vectors}. The correspondence between the annotation vectors and image regions is because convolution represents moving, overlapping, and localized dot-products and pooling results in effectively larger receptive fields. Therefore each depth vector in the output volume corresponds to a $d$-dimensional encoding of a localized image region. The set of all annotation vectors produced by the DenseNet is represented by
\begin{equation}
    A = \{\mathbf{a}_1 , \mathbf{a}_2 ,\dots,  \mathbf{a}_L\}
    \label{eq:3}
\end{equation}
where $\mathbf{a}_l  \in \mathbb{R}^d$ and number of regions is given by $L=hw$.

\subsection{Decoder: Gated Recurrent Units with Contextual Attention Localization}
\begin{figure}[!b]
    \centering
    \includegraphics[width=.3\linewidth]{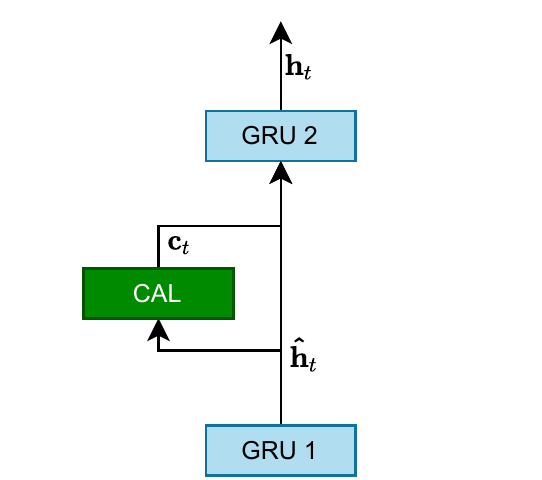}
    \caption{The decoder contains a contextual attention localization (CAL) unit between two gated recurrent units (GRUs).}
    \label{fig:decoder}
\end{figure}

When an encoder extracts high-level visual features from input images, a Gated Recurrent Unit (GRU) \cite{Chung2014} can be used to generate a variable-length text sequence character by character, conditioned on the previous output character $\mathbf{y}_{t-1}$, the current hidden state $\mathbf{h}_t$ and a context vector $\mathbf{c}_t$. Our decoder consists of a Contextual Attention Localization (CAL) unit sandwiched between two GRUs as shown in Figure \ref{fig:decoder}.

\subsubsection{First GRU}
The first GRU is used to compute a predicted hidden state $\mathbf{\hat{h}}_t$ via the following recurrent equations
\begin{align}
    \mathbf{z}_{1,t}&=\sigma(\mathbf{W}_{yz}\mathbf{E}\mathbf{y}_{t-1} + \mathbf{U}_{hz}\mathbf{h}_{t-1})
    \label{eq:z1t}
    \\
    \mathbf{r}_{1,t}&=\sigma(\mathbf{W}_{yr}\mathbf{E}\mathbf{y}_{t-1} + \mathbf{U}_{hr}\mathbf{h}_{t-1})
    \label{eq:r1t}
    \\
    \mathbf{\widetilde{h}}_t&=\tanh(\mathbf{W}_{yh}\mathbf{E}\mathbf{y}_{t-1} + \mathbf{r}_{1,t} \otimes (\mathbf{U}_{rh}\mathbf{h}_{t-1}))
    \label{eq:htilde}
    \\
    \mathbf{\hat{h}}_t&= \mathbf{z}_{1,t} \otimes \mathbf{h}_{t-1} + (1-\mathbf{z}_{1,t}) \otimes \mathbf{\widetilde{h}}_{t}
    \label{eq:hhat}
\end{align}
where $\sigma$ is the logistic sigmoid function, $\otimes$ is a point wise multiplication, $\mathbf{E}\in\mathbb{R}^{m\times k}$ is a learnable lower-dimensional projection matrix for getting rid of the wasteful $1$-of-$k$ coding of character vector $\mathbf{y}_{t-1}$. All other matrices within each equation represent learnable transformations into a constant dimensional space. They can be divided into character embedding transformations $\{\mathbf{W}_{yz},\mathbf{W}_{yr},\mathbf{W}_{yh}\}\in \mathbb{R}^{h \times m}$ and recurrent transformations $\{\mathbf{U}_{hz},\mathbf{U}_{hr},\mathbf{U}_{rh}\}  \in \mathbb{R}^{h \times h}$. The recurrent hidden state $\mathbf{h}_{t-1}$ from the previous time step is computed using Equation \eqref{eq:ht}. The predicted hidden state $\mathbf{\hat{h}}_t$ is then used in the dynamic computation of the attention weights as described next.

\subsubsection{Contextual Attention}
Attention has been successfully used in computer vision \cite{Bahdanau2016} and also in handwriting recognition \cite{zhang2017, zhang2018multi}. The benefit of using attention is that it adds context to the decoder. For example, an image region should be classified as a dog by looking at the dog and not by looking at its owner. While a dog can appear anywhere in an image, characters in a text sequence cannot appear at arbitrary locations. Text has a strong structure and therefore \emph{attention itself needs context}. In simpler words, where to focus next depends on where the model has focused previously.

Ideally, we would like our text recognizer to \emph{read} like we do. That is, it should focus on the relevant image region when recognizing each character or word. However, for images containing text, multiple instances of a character or word can appear at multiple locations. Nothing stops an attention model from re-attending a previously attended location or from giving the right answer by attending the wrong location. Therefore, \emph{for text}, the decision for attention needs to depend on the history of attention. In our model, attention at time $t$ is made to depend on previous values of attention. This is achieved through a coverage vector, that represents a history of attention already given to each location.

\begin{figure}[!t]
    \centering
    \includegraphics[width=.9\linewidth]{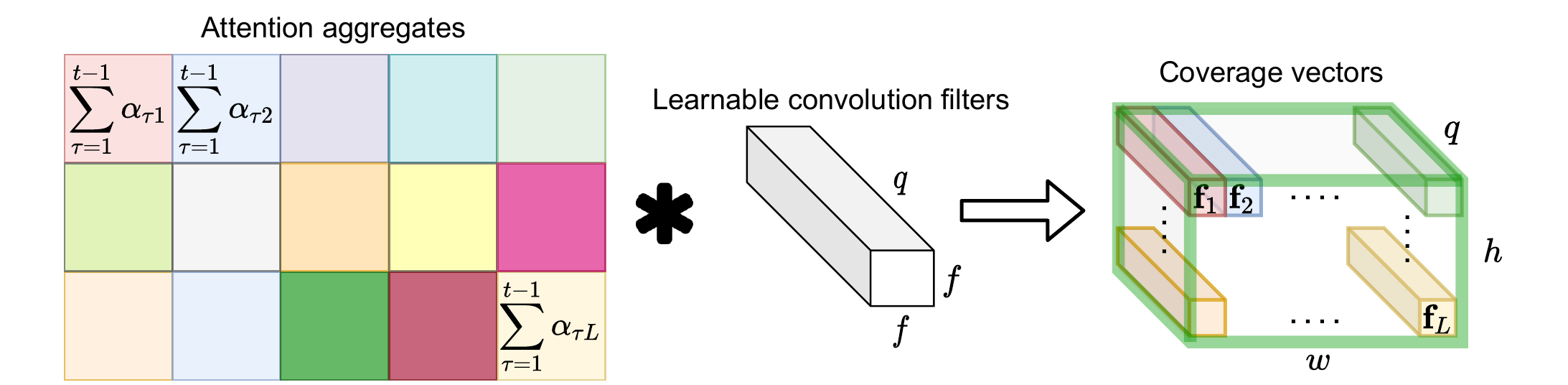}
    \caption{Computation of coverage vector.}
    \label{fig:coverage}
\end{figure}

Let attention weight $\alpha_{tl}$ denote the importance of image region $l$ at time $t$. By aggregating the attentions through time, we can compute the $h\times w$ aggregated attention array $S^\alpha$ at time $t$ via
\begin{align}
    S^\alpha_{tl} &= \sum_{\tau=1}^{t-1} \alpha_{\tau l}
    \label{eq:S_alpha_tl}
\end{align}
The 2-dimensional array $S^\alpha$ can be convolved with $q$ filters of size $f\times f$ arranged as a volume $Q$ to obtain an $h\times w\times q$ volume $F_t$ as
\begin{align}
    F_t &= S^\alpha \ast Q
    \label{eq:coverage_volume}
\end{align}
Features in $F_t$ can be interpreted as the \emph{history of attention} until time $t$ and $Q$ is a learnable layer of convolution filters. Similar to how annotation vector $\mathbf{a}_l$ is formed, the coverage vector $\mathbf{f}_l$ is the $q$-dimensional vector at location $l$ of volume $F_t$ as shown in Figure \ref{fig:coverage}. Each vector $\mathbf{f}_l$ can be interpreted as the \emph{coverage} given to location $l$ until time $t$. Therefore, $\mathbf{f}_l$ is called a \emph{coverage vector}. The coverage vectors describe the attention given \emph{so far} to different regions in a \emph{dynamic} fashion. In contrast, the annotation vectors $\mathbf{a}_1,\dots,\mathbf{a}_L$ describe input image regions in a \emph{static} fashion.

The dynamic attention values $\alpha_{tl}$ for the current time step $t$ can then be computed via
\begin{align} 
    e_{tl} & =\mathbf{v}_{a}^{T}\tanh(\mathbf{W}_a\mathbf{\hat{h}}_{t-1} + \mathbf{U}_{a}\mathbf{a}_{l} + \mathbf{U}_{f}\mathbf{f}_{l})
    \label{eq:eq_etl}
    \\
    \alpha_{tl} &= \frac{\exp(e_{tl})}{\sum_{u=1}^{L}\exp(e_{tu})}
    \label{eq:eq_atl} 
\end{align}
where $\mathbf{v}_a \in \mathbb{R}^{n}$, $\mathbf{W}_a \in \mathbb{R}^{n \times h}$, $\mathbf{U}_a \in \mathbb{R}^{n \times d}$, and $\mathbf{U}_f \in \mathbb{R}^{n \times q}$ are all learnable projection parameters. Notice that the softmax operation in Equation \eqref{eq:eq_atl} ensures that attention weights $\alpha_{tl}$ at time $t$ represent probabilities.

\begin{figure}[!b]
    \centering
    \includegraphics[width=.6\linewidth]{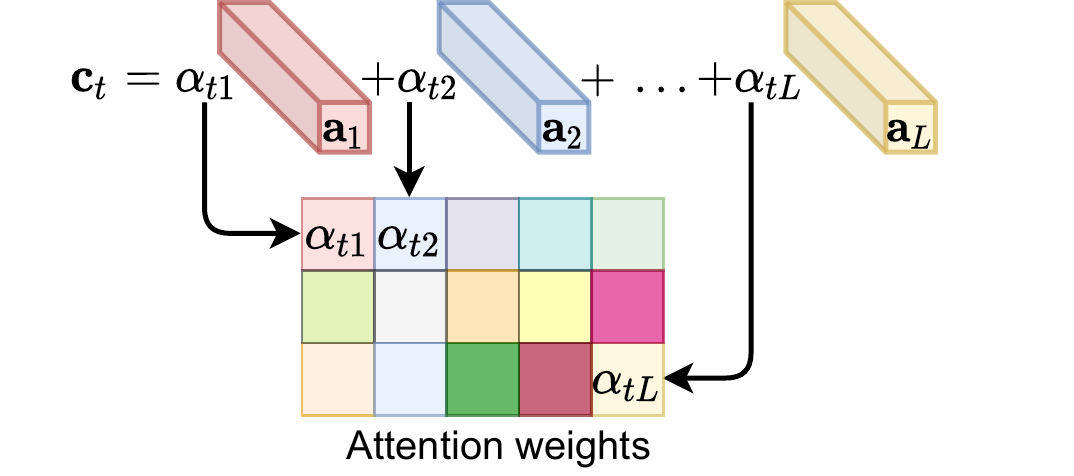}
    \caption{Attention weight $\alpha_{tl}$ determines importance of location $l$ in determining context vector $\mathbf{c}_t$ at time $t$.}
\label{fig:attention_weights}
\end{figure}

A \emph{dynamic} representation of the relevant part of the input image at time $t$ can now be defined. As illustrated in Figure \ref{fig:attention_weights}, it is computed as the expectation of the annotation vectors
\begin{equation}
    \mathbf{c}_t=\sum_{l=1}^{L}\alpha_{tl}\mathbf{a}_l
    \label{eq:eq_ctx}
\end{equation}
where attention weights $\alpha_{tl}$ determine the importance of each image region at time $t$. They determine the image \emph{context} in which the decision for output character at time $t$ will be made. Accordingly, the expectation of the annotation vectors can be termed as the \emph{context vector} $\mathbf{c}_t$.

\subsubsection{Second GRU}
The second GRU computes the final hidden state $\mathbf{h}_t$ using predicted hidden state $\mathbf{\hat{h}}_t$ and the context vector $\mathbf{c}_t$ via the following recurrent equations
\begin{align}
    \mathbf{z}_{2,t}&=\sigma(\mathbf{U}'_{hz}\mathbf{\hat{h}}_{t} + \mathbf{C}_{cz}\mathbf{c}_{t})
    \label{eq:z2t}
    \\
    \mathbf{r}_{2,t}&=\sigma(\mathbf{U}'_{hr}\mathbf{\hat{h}}_{t} + \mathbf{C}_{cr}\mathbf{c}_{t})
    \label{eq:r2t}
    \\
    \mathbf{\widetilde{h}}_t&=\tanh(\mathbf{r}_{2,t} \otimes (\mathbf{U}'_{rh}\mathbf{\hat{h}}_{t}) + \mathbf{C}_{ch}\mathbf{c}_{t} )
    \label{eq:htildet}
    \\
    \mathbf{h}_t&= \mathbf{z}_{2,t} \otimes \mathbf{\hat{h}}_{t} + (1-\mathbf{z}_{2,t}) \otimes \mathbf{\widetilde{h}}_{t}
    \label{eq:ht}
\end{align}
where learnable matrices can be divided into recurrent transformations $\mathbf{U}'_{hz},\mathbf{U}'_{hr},\mathbf{U}'_{rh}\}  \in \mathbb{R}^{h \times h}$, and context transformations $\{\mathbf{C}_{cz},\mathbf{C}_{cr},\mathbf{C}_{ch}\} \in \mathbb{R}^{h \times d}$.

The conditional probabilities of the next character are computed as
\begin{equation}
p(\mathbf{y}_t | \mathbf{y}_{t-1}; \mathbf{X} ) = \mathcal{S}(\mathbf{W}_{o}(\mathbf{W}_{y}\mathbf{E}\mathbf{y}_{t-1} + \mathbf{W}_h\mathbf{h}_{t} + \mathbf{W}_c\mathbf{c}_{t}))
\label{eq_condprob}
\end{equation}
where $\mathcal{S}(\cdot)$ is the softmax activation function and $\mathbf{W}_{o} \in \mathbb{R}^{k \times m}$, $\mathbf{W}_{h} \in \mathbb{R}^{m \times h}$ and $\mathbf{W}_{c} \in \mathbb{R}^{m \times d}$ are all learnable projection parameters.

The sequence of probability vectors is converted into an optimal sequence of output characters $\mathbf{y}_1, \mathbf{y}_2, \dots, \mathbf{y}_T$ via beam search \cite{wiseman_2016_beamsearch} whereby only the $b$ most-probable branches are pursued at each time step.

\subsection{Attention Localization}
Text is made up of very localized image regions so that a single character appears in a small contiguous area. It cannot appear scattered all over the image. Therefore, in addition to context of attention, text requires \emph{localization of attention} as well. 

This behaviour can be induced by encouraging the attention weights $\alpha_{tl}$ to be sparse. Accordingly, we define the attention localization penalty as
\begin{equation}
    \mathcal{L}(\theta)=\sum_{t=1}^T \sum_{l=1}^L \vert\alpha_{tl}\vert
    \label{eq:att_loc}
\end{equation}
which is the $\ell_1$-norm of the vector of attention values. Minimizing $\ell_1$-norm encourages only few $\alpha_{tl}$ to be large and all the rest to be close to zero. This discourages the model from attending to multiple locations at the same time.

\subsection{Error Function}
The proposed model is trained jointly for recognition of offline handwritten text. In the model, the next character is predicted depending on the previous character and input image as explained in Equation  \eqref{eq_condprob}. The objective of the training is to maximize the probability of the predicted character. In order to train the model to classify using contextual attention, regularized cross-entropy is used as the error function
\begin{equation}
    \mathcal{C}_n(\theta)=-\sum_{t=1}^{T_n} \ln p(\mathbf{y}_n^{(t)}|\mathbf{d}_n^{(t)}, \mathbf{x}_n) + \lambda r(\theta)
    \label{eq:cross_entropy}
\end{equation}
where $\mathbf{y}_n^{(t)}$ is the predicted character at timestamp $t$ of the ${n}^{th}$ training image, $r(\theta)$ is the $\ell_2$-norm of all weights except those from the convolution layers and $\lambda>0$ is the regularization hyperparameter.

In order to train the model to classify using \emph{localized contextual attention}, cross-entropy can be augmented with our attention localization penalty \eqref{eq:att_loc} to yield our proposed error function
\begin{equation}
    \mathcal{E}_n(\theta)=\mathcal{C}_n(\theta)+\gamma \mathcal{L}_n(\theta)
    \label{eq:overall_error_function}
\end{equation}
where hyperparameter $\gamma>0$ controls the tradeoff between classification and localization.

\section{Datasets}
\label{sec:datasets}
We evaluate our proposed method on handwritten Urdu as well as Arabic.

\subsection{Urdu Dataset}
\begin{figure}[!t]
    \centering
    \resizebox{0.5\textwidth}{!}{
    \centering
    \begin{tabular}{|c|c|c|c|}
        \hline
        Blank paper & 
        \includegraphics[width=.24\linewidth]{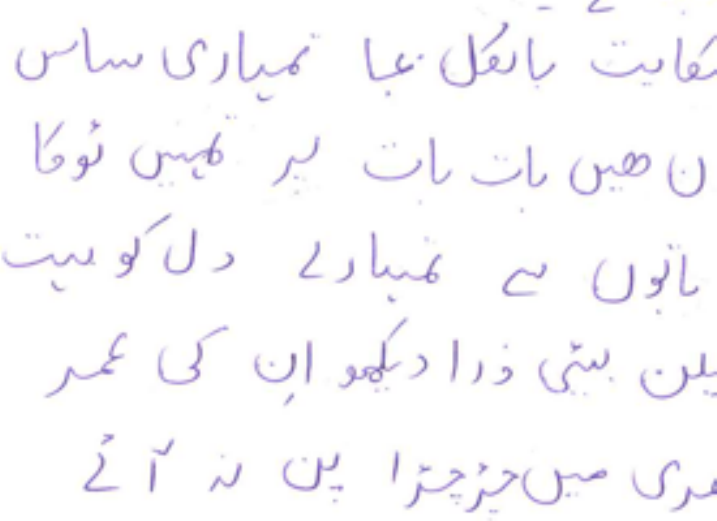} & \includegraphics[width=.24\linewidth]{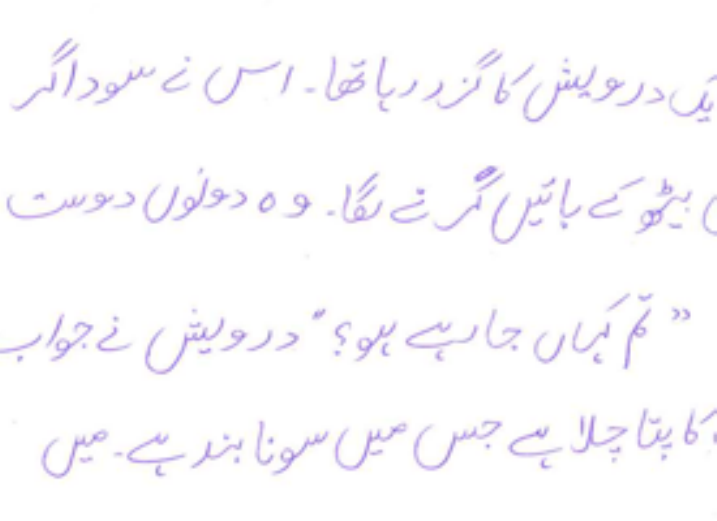} & \includegraphics[width=.24\linewidth]{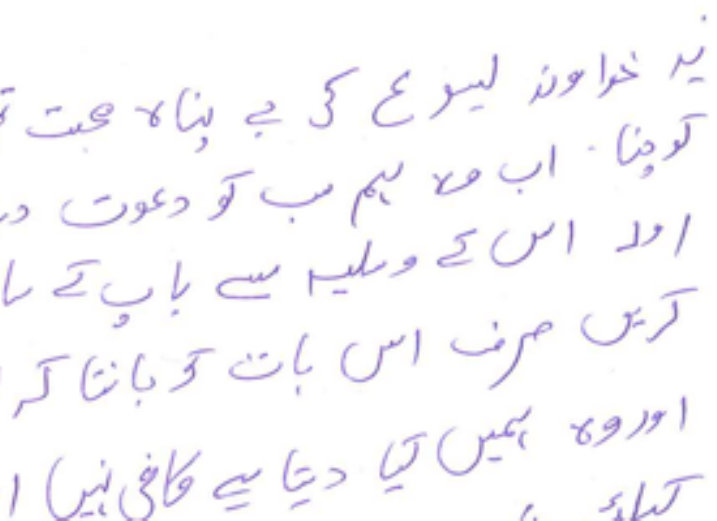} \\
        \hline
        \makecell{Ruled paper} &  
        \includegraphics[width=.24\linewidth]{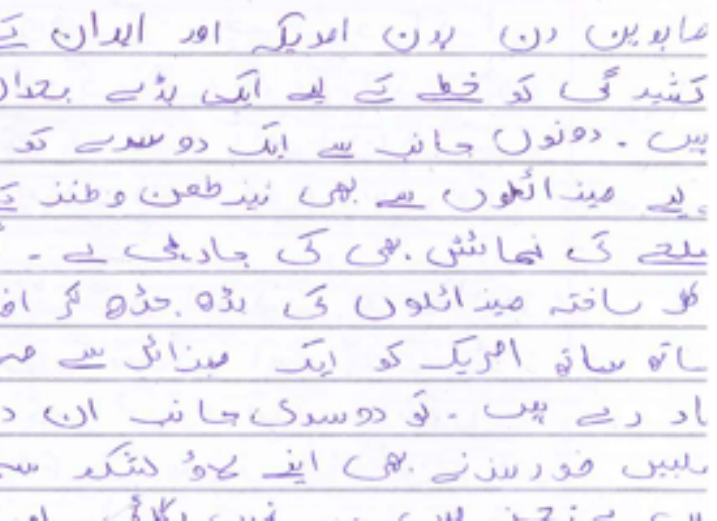} & \includegraphics[width=.24\linewidth]{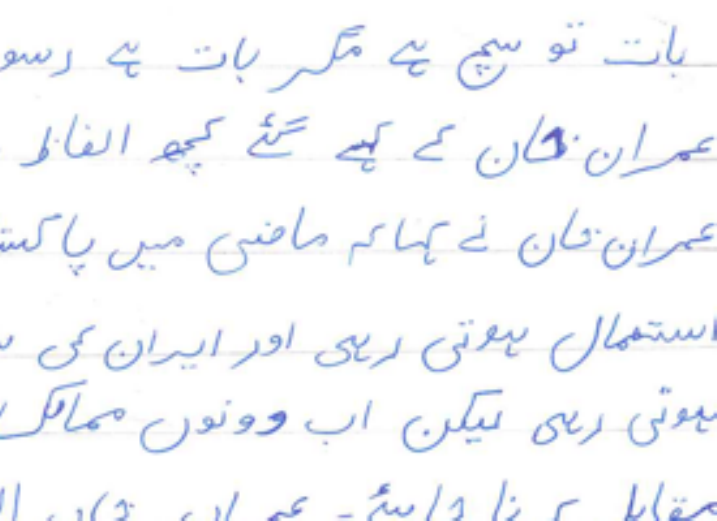} & \includegraphics[width=.24\linewidth]{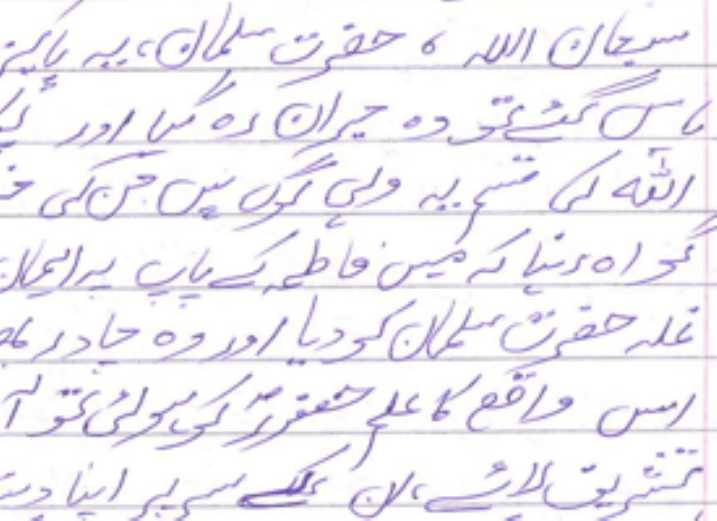} \\
        \hline
        \makecell{Background\\colour} &  \includegraphics[width=.24\linewidth]{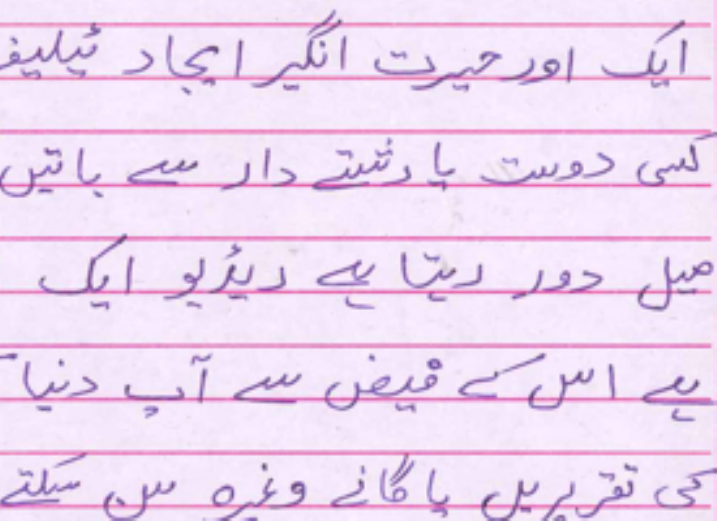} & \includegraphics[width=.24\linewidth]{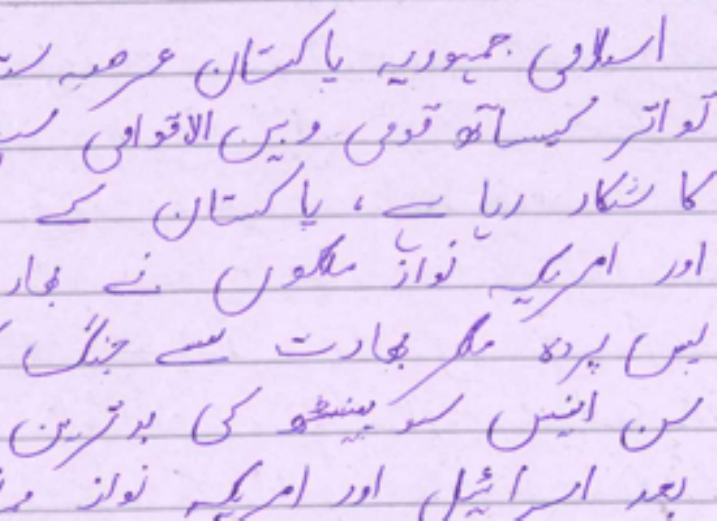} & \includegraphics[width=.24\linewidth]{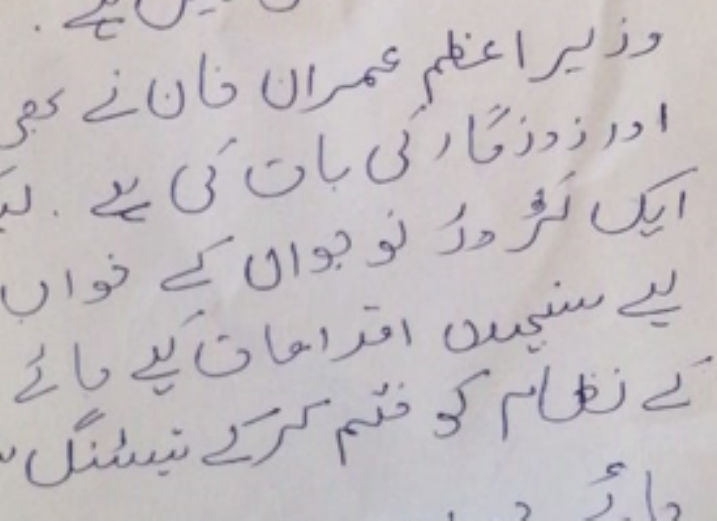} \\
        \hline
        \makecell{Ink colour} &  \includegraphics[width=.24\linewidth]{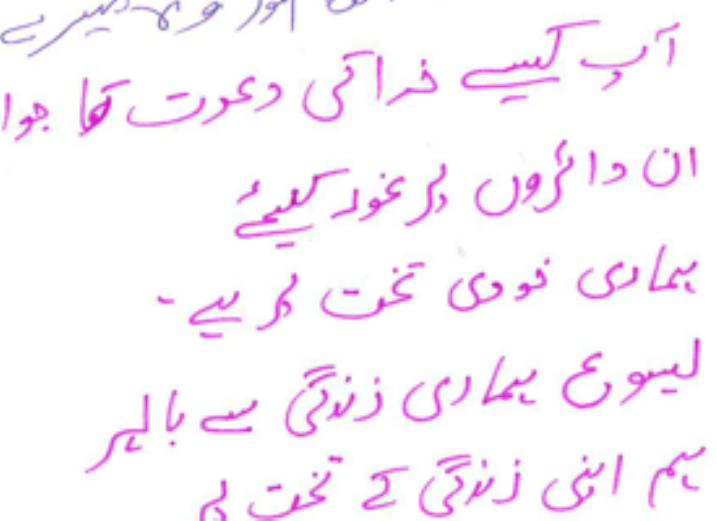} & \includegraphics[width=.24\linewidth]{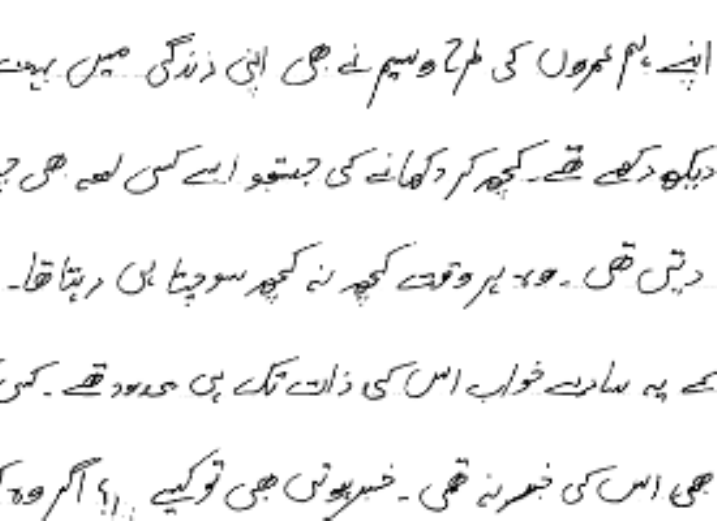} & \includegraphics[width=.24\linewidth]{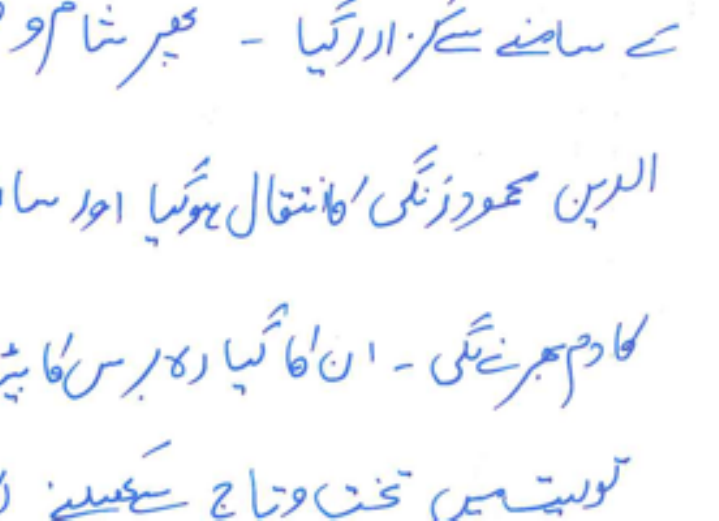} \\   
        \hline
        \makecell{Pen type} &  
        \includegraphics[width=.24\linewidth]{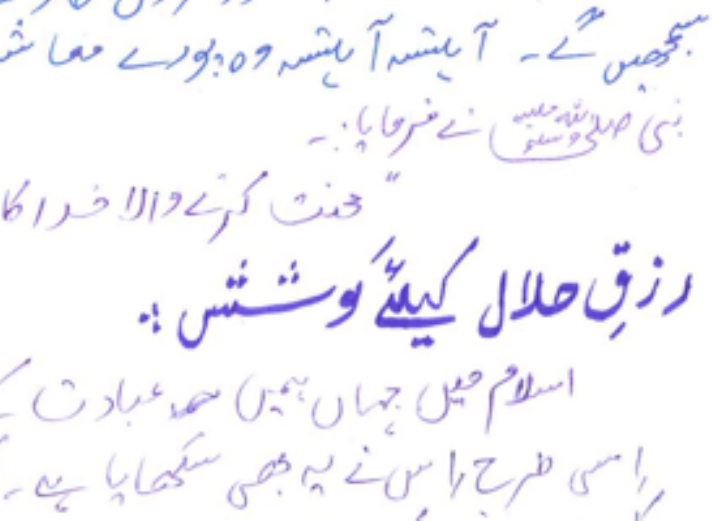} & \includegraphics[width=.24\linewidth]{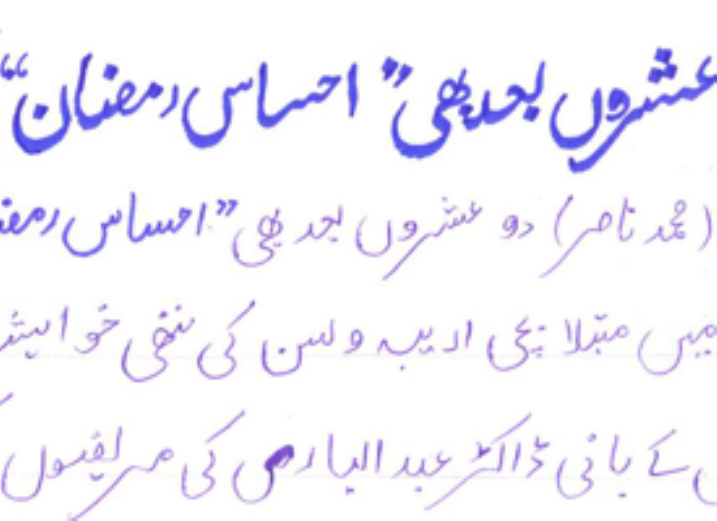} & \includegraphics[width=.24\linewidth]{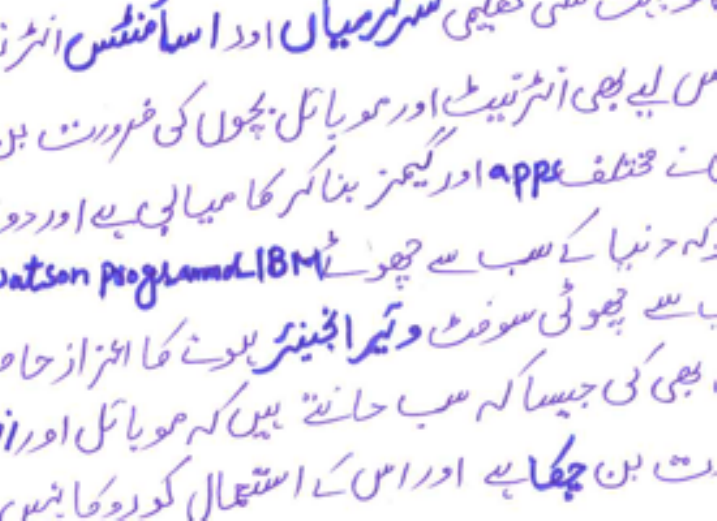} \\   
        \hline
    \end{tabular}
    }
    \caption{Sample page images from the PUCIT-OHUL dataset demonstrating challenges related to different paper types, backgrounds, colours, ink types and pen types.}
    \label{fig:pucit_ohul_smaples}
\end{figure}

The PUCIT-OHUL dataset introduced in \cite{anjum_icfhr2020}
is a multi-writing style dataset with different pen types, ink types, text sizes, and background types and colours. It was collected using $100$ students between $20$ and $24$ years of age. A total of $479$ pages of text were collected. Each page was scanned at $200$ DPI and text lines were manually segmented. The line images were not deskewed. As shown in Figure \ref{fig:pucit_ohul_smaples}, the dataset contains different types of page backgrounds, page colors, ink types and pen types. An overview of the subject categories covered in the dataset is shown in Table \ref{tab:subjects}. The writers were not given any instructions regarding what subject to choose and were completely free in their choices. The detailed vocabulary of unique characters and symbols extracted from the dataset is given in Table \ref{tab:dataset_vocabulary}.

Unsurprisingly, labeling errors in datasets adversely impact training as well as error rates \cite{aradillas_icfhr2020_dataset_purging}. Accordingly, we have re-annotated the complete PUCIT-OHUL dataset. The key outcomes of our re-annotation process are as follows.
\begin{enumerate}
    \item We have added $92$ new lines to the original test set.
    \item In an attempt to reduce annotator bias, $7$ annotators were used.
    \item Vocabulary size was increased from $98$ in the original ground-truth to $130$ in our updated version. We annotated $31$ additional characters from the English alphabet and $1$ from Urdu that had been used by the original writers but the original ground-truth did not include them. In the updated annotations, we label each and every character.
    \item For a few cases, there was a mismatch between images and annotations. Such mismatches have been fixed.
    \item In the original ground-truth, spelling mistakes in the handwritten text were sometimes \emph{corrected} by the annotators. In our re-annotations, such mistakes have been recorded as they appear with no attempt to correct them.
    \item Original ground-truth had a non-trivial amount of errors in the annotation of diacritics. Since diacritics are usually small and the mind can \emph{auto-correct} mistakes in diacritics, such errors are easy to overlook. A lot of attention was focused on correct re-annotation of diacritics.
    \item We categorize crossed-out text as either readable or non-readable. Such information can be used to \emph{not penalize} a recognizer that correctly recognizes readable crossed-out text.
\end{enumerate}

\begin{table}
	\caption{Categorization of the content in the PUCIT-OHUL dataset in terms of subjects, references, pages, and lines.}
	\centering
	\begin{tabular}{|c|c|c|c|c|}
		\hline 
		\bf{} & \bf{Subject} & \bf{References}  & \bf{Pages} & \bf{Lines}
		\\ 
		\hline
		1 &  Art & 1 & 5 & 72
		\\
	    	\hline
		2 & World & 11 & 52 & 805
		\\
			\hline
		3 &  Economy & 3 & 15 & 277
		\\
			\hline
		4 &  Social & 19 & 84 & 1291
		\\
			\hline
		5 &  Literature & 17 & 83 & 1293
		\\
			\hline
		6 &  Wildlife & 1 & 5 & 106
		\\
			\hline
		7 &  Sports & 4 & 21 & 317
		\\
			\hline
		8 &  Health & 1 & 5 & 77
		\\
			\hline
		9 &  Religion & 8 & 40 & 662
		\\
			\hline
		10 &  Nature & 2 & 9 & 129
		\\
				\hline
		11 &  Science & 3 & 14 & 203
		\\
			\hline
		12 &  Politics & 14 & 67 & 1005
		\\
			\hline
		13 &  History & 10 & 52 & 683
		\\
			\hline
		14 &  Education & 6 & 27 & 479
		\\
			\hline
		 &  Total & 100 & 479 & 7,399
		\\
		\hline 
	\end{tabular}
	\label{tab:subjects} 
\end{table}

\begin{table}%[!htt]
	\caption{Vocabulary extracted from the PUCIT-OHUL dataset.}
	\centering
	\includegraphics[width=.9\linewidth]{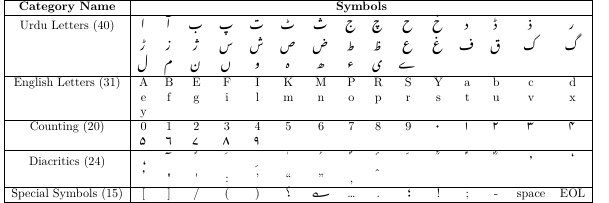}
	\label{tab:dataset_vocabulary} 
\end{table}

\begin{table}[!h]%[!htt]
	\caption{Comparison with existing offline handwritten Urdu text datasets}
	\centering
	%\resizebox{0.5\textwidth}{!}{
	%\centering
	\begin{tabular}{|c|c|c|c|c|c|c|c|c|c|}
		\hline 
		\bf{Dataset} & \bf{\makecell{Total\\ Lines}} & \bf{\makecell{Total\\ Words}}&\bf{\makecell{Total\\ characters}}&\bf{\makecell{Total\\ Writers}}&\bf{\makecell{Vocabulary\\ Size}}\\

		\hline
		%&&&&&
		%\\
		UCOM/UNHD \cite{Ahmed2019} & 10,000 & 312,000& 1,872,000& 500&59
		\\ 
		CENIP-UCCP \cite{Raza2012} & 2,051 &23,833 &-&  200&-\\
		
		Custom dataset \cite{Hassan2019} & 6,000 &86,400&432,000 &600&-
		\\ 
		PUCIT-OHUL & $\textbf{7,401}$  &$\textbf{80}$,$\textbf{059}$& $\textbf{367}$,$\textbf{925}$&$\textbf{100}$&$\textbf{130}$
		\\ 
	%&&&&&
	%	\\
		\hline 
	\end{tabular}
	%}
	\label{tab:dataset_comparison} 
\end{table}

Table \ref{tab:dataset_comparison} compares the PUCIT-OHUL dataset with existing datasets. Existing datasets are problematic for two reasons. Firstly, sometimes data statistics are not reported correctly. For instance, UCOM/UNHD \cite{Ahmed2019} claims to have $10,000$ text lines, but only $700$ of those lines are unique in terms of semantic content. Secondly, all $3$ datasets in \cite{Ahmed2019, Raza2012, Hassan2019}  are either publicly unavailable or only partially available. In contrast, the PUCIT-OHUL dataset with our updated annotations is publicly available\footnote{\url{http://faculty.pucit.edu.pk/nazarkhan/work/urdu_ohtr/pucit_ohul_dataset.html}} in its entirety.

The dataset is divided randomly into training, validation, and testing sets with respect to writers. Out of the $100$ writers, we randomly selected $75$ writers for the training set, $13$ for the validation set and the remaining $12$ for the testing set. In terms of lines, $5580~(75.42\%)$ of the lines were included in the training set, $907~(12.33\%)$ in the validation set, and $912~(12.25\%)$ in the testing set.

\subsection{Arabic Dataset}
The KFUPM Handwritten Arabic TexT (KHATT) \cite{Mahmoud2014} dataset is an unconstrained offline handwritten Arabic dataset. We use the default split of the unique text portion of the dataset which consists of $4,825$ training, $937$ validation, and $966$ testing line images. The dataset contains $63$ unique symbols.

\section{Experiments and Results}
\label{sec:exp_results}
\subsection{Experimental Setup}
The proposed model is composed of a DenseNet encoder and a contextual attention mechanism sandwiched between two GRU decoders. Dimensions of all learnable matrices and vectors in our decoding framework are given in Table \ref{tab:dimensions}.

\begin{table}[h!]
    \centering
    \caption{Dimensions of learnable matrices and vectors in the proposed decoding framework.}
    \begin{tabular}{ll}
        \textbf{Equation} & \textbf{Dimensions}
        \\\hline
        \eqref{eq:z1t}, \eqref{eq:r1t}, \eqref{eq:htilde} &  $k=130, m=256, h=256$\\
        \eqref{eq:coverage_volume} & $f=11, q=512$\\
        \eqref{eq:eq_etl} & $n=256, d=684$\\
    \end{tabular}
    \label{tab:dimensions}
\end{table}

To reduce overfitting, $4\%$ salt and pepper noise is added to training images with probability of salt being $20\%$ and probability of pepper being $80\%$. During training, we use batch normalization and dropout. Dropout is only used at convolution layers with $20\%$ drop ratio.  We also apply regularization with $\lambda=1e^{-4}$ in the cross-entropy loss \eqref{eq:cross_entropy} and $\gamma=1$ in the overall error function \eqref{eq:overall_error_function}. We use an Adadelta optimizer with gradient clipping for better optimization. All training and testing images are resized to $100 \times 800$ pixels. Before resizing, if the width is less than $300$, we first double the width of the image by extra white padding and then resize the result to $100 \times 800$. This way we avoid excessive stretching in images with small width.

The proposed model is trained and tested on Urdu PUCIT-OHUL \cite{anjum_icfhr2020} as well as Arabic KHATT \cite{Mahmoud2014} datasets.

\subsection{Quantitative Results}
We use character and word error rates to quantitatively compare the proposed method with other approaches. The character error rate (CER) between target and output text is calculated using the Levenshtein edit distance formula
\begin{equation}
\text{CER} = \frac{\text{I} + \text{S} + \text{D}}{\text{N}} \times 100
\label{eq:cer}
\end{equation}
where I, S and D represent the minimum numbers of insertions, substitutions and deletions required to convert the target text string into the output text string and N represents the total number of characters in the target text\footnote{Theoretically, CER can be greater than $100$ but for any reasonably performing recognizer, it is typically less than $100$.}. To calculate word error rate (WER) Equation \eqref{eq:cer} can be used after replacing characters with words. To calculate accuracies, both error rates are subtracted from $100$.

\begin{table}[!t]
	\caption{Comparison of different models for offline handwritten Urdu text recognition on PUCIT-OHUL dataset.}
	\centering
    %\resizebox{0.5\textwidth}{!}{
	%\centering
	\begin{tabular}{|c|c|c|}
		\hline
 \bf{\makecell{Models}}  & \bf{\makecell{Character Level\\ Accuracy}}&\bf{\makecell{Word Level\\ Accuracy}} \\
		\hline 
			CNN+GRU &  $33.00$ & $22.44$\\\hline
			CNN+LSTM & $32.04$ & $21.42$ \\\hline
			CNN+BGRU & $32.76$ & $23.16$\\\hline
			CNN+BLSTM \cite{Hassan2019} & $32.02$ & $22.23$
		    \\\hline
            Contextual Attention \cite{anjum_icfhr2020} & $77.74$ & $45.55$
            \\\hline
            \makecell{Contextual Attention\\Localization} & $\mathbf{79.76}$ & $\mathbf{47.93}$
            \\\hline 
	\end{tabular}
%}
	\label{tab:comparison_urdu} 
\end{table}

\subsubsection{PUCIT-OHUL dataset}
Table \ref{tab:comparison_urdu} shows the comparison of previous models with our proposed model. We compare our model with CNN encoder based uni-directional as well as bidirectional decoders that do not use attention. The CNN architecture that we use has previously been used for handwritten Urdu, Arabic, English and French. The CNN model used in the first $4$ models at Table \ref{tab:comparison_urdu} is based on the architecture from \cite{Hassan2019} and each method was trained on our dataset. In our experiments, bidirectional decoders were able to improve character level accuracies by less than $2\%$. Since character level accuracies were so low, word level accuracies are understandably poor. This is why previous work on handwritten Arabic-like scripts does not report word level accuracies. In contrast, our DenseNet encoder jointly trained with an attention based decoder almost doubles the character level accuracy and word level accuracy. We believe the reasons for such drastic improvement in results are that: The DenseNet encoder learns more diverse features compared to the CNN model of previous approaches. In addition, attention reduces the need for bidirectional decoding. Finally, using the concept of \emph{coverage of attention} captures the right-to-left nature of Urdu.

Results show that incorporation of the DenseNet encoder with attention based decoding almost doubles the character level accuracy and word level accuracy. Furthermore, it can be observed that by using localized attention accuracy furthers increases by a value of around $2\%$.

The proposed recognizer based LSTM along with DenseNet significantly improves the character and word level accuracy as compared to already existing techniques in literature for Urdu text recognition. The use of attention helps the model in looking at the appropriate location, and attention localization further helps to localize it. This greatly benefits character/word level accuracy.

\subsubsection{KHATT dataset}
We also compare proposed models with the MDLSTM model used for recognition of Arabic offline handwritten KHATT dataset in Table \ref{tab:comparison_arabic}. We use KHATT unique text lines dataset. We don't use any skew and slant correction. Instead of a multidimensional decoder, an attention based uni-dimensional encoder without any specific pre-processing steps gives almost equal results as achieved by \cite{Ahmed2017}. Adding localization further increases the accuracy by almost $2\%$.

\begin{table}[!b]
	\caption{Comparison of proposed models with models proposed for recognition of offline handwritten Arabic KHATT dataset.
	}
	\centering
	%\resizebox{0.5\textwidth}{!}{
	%\centering
	\begin{tabular}{|c|c|c|c|}
		\hline 
	{\bf{Models}}& {\bf{Pre-processing}} &
\bf{\makecell{Character Level\\ Accuracy}} &\bf{\makecell{Word Level\\ Accuracy}} \\
		\hline 
		    \makecell{MDLSTM \cite{Ahmed2017}} & \makecell{Skew, slant \\
		    correction,
pruning,\\
and normalization}& $75.80$&$-$\\\hline

            \makecell{Contextual Attention \cite{anjum_icfhr2020}} & Normalization & $75.16$ & $33.88$
            \\\hline
             
            \makecell{Contextual Attention\\ Localization} & Normalization & $\mathbf{77.15}$ & $\mathbf{35.83}$
            \\\hline
	\end{tabular}
	%}
	\label{tab:comparison_arabic} 
\end{table}

\subsection{Qualitative}
\begin{figure}[!t]%[!htt]
    \centering
    \captionsetup[subfigure]{labelformat=empty}
    \subfloat[CER=$0\%$]{
    \includegraphics[width=.7\linewidth]{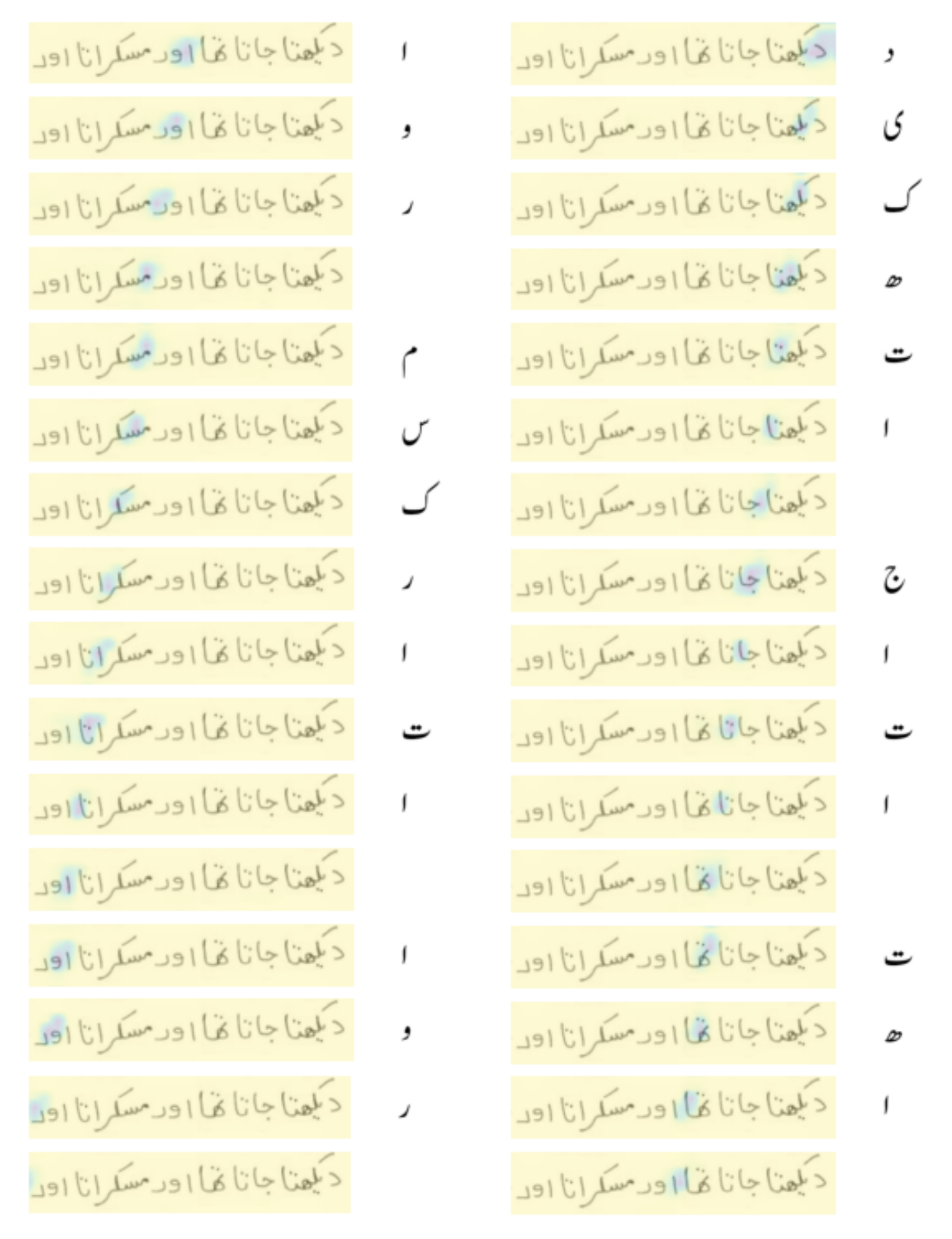}\label{l1}
    }
    \caption{Attention visualization. Our model implicitly segments and recognizes every character by focusing only on localized, relevant areas. It also learns to focus in context.}
    \label{fig:attvis}
\end{figure}

In order to explain the results of the model, we can visualize the attention weights $\alpha_{tl}$ for region $l$ at time $t$. Let $\alpha_t$ denote the attention map array for all locations at time $t$. This attention map can be superimposed on the original image. Figure \ref{fig:attvis} demonstrates the regions of attention for the proposed CALText model when decoding a sequence of output characters. It can be seen that the model implicitly segments and recognizes each character by focusing only on localized, relevant areas while considering the context of previously attended locations. The purple region indicates the region of attention. It can be observed that the model has learned to attend from right-to-left which is the natural order of Urdu script. The most probable character at each location inferred after beam-search is shown on the right.

\begin{figure*}
    \centering
    \begin{tabular}{cc}
        \textbf{Contextual Attention} & \textbf{Contextual Attention Localization}
        \\
        \includegraphics[width=.4\linewidth]{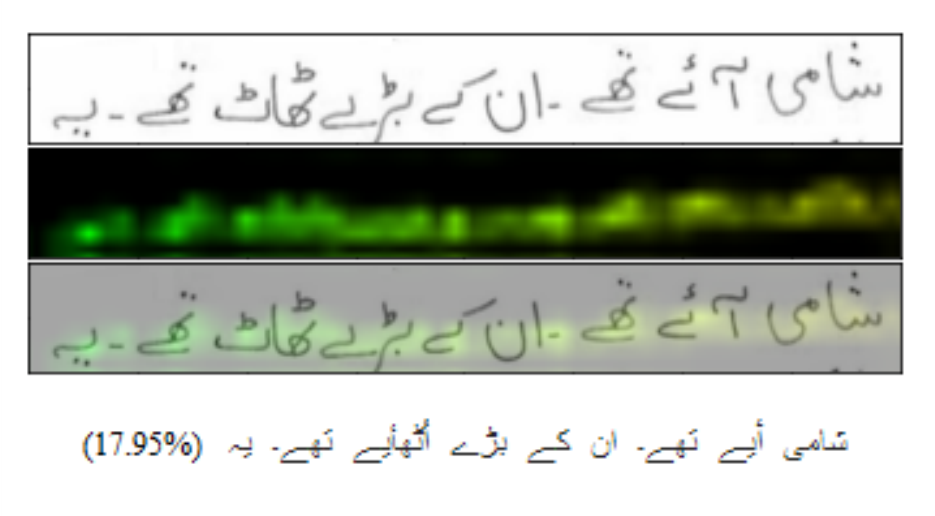} & \includegraphics[width=.4\linewidth]{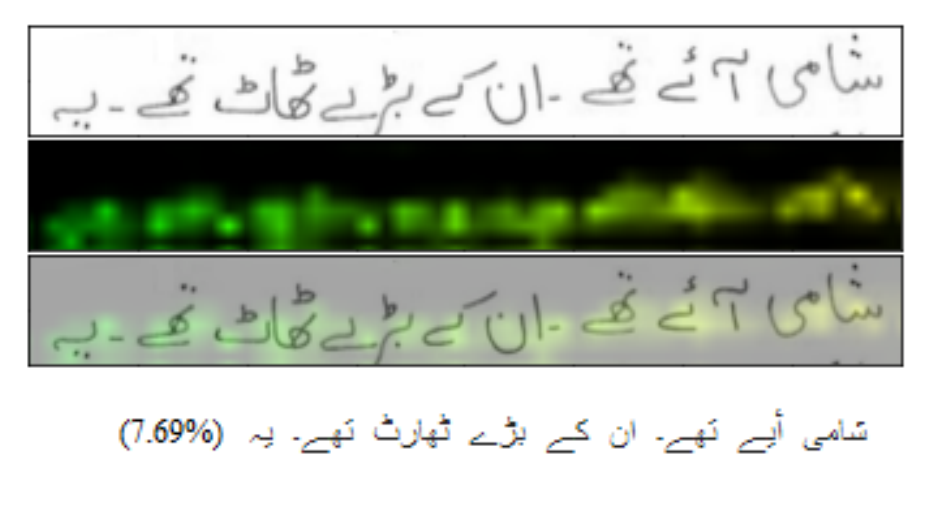}
        \\
        \includegraphics[width=.4\linewidth]{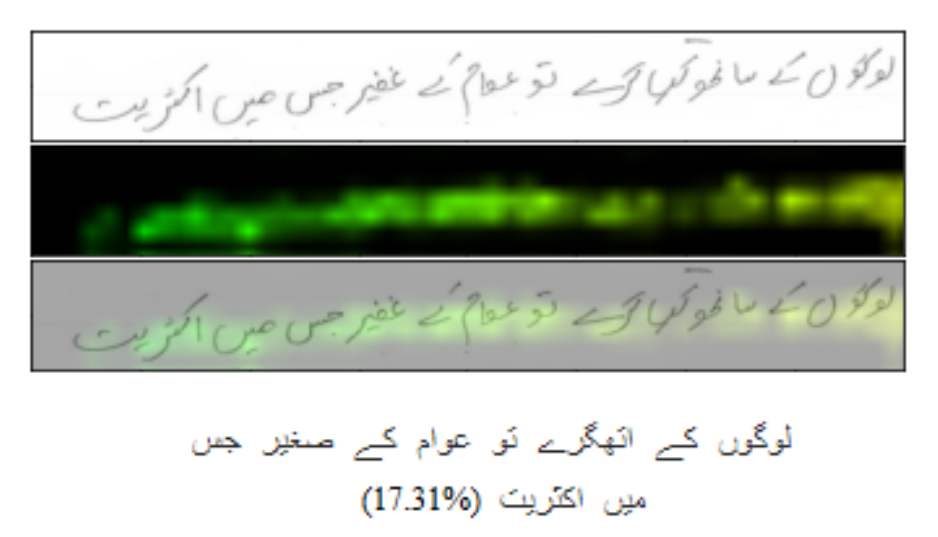} & \includegraphics[width=.4\linewidth]{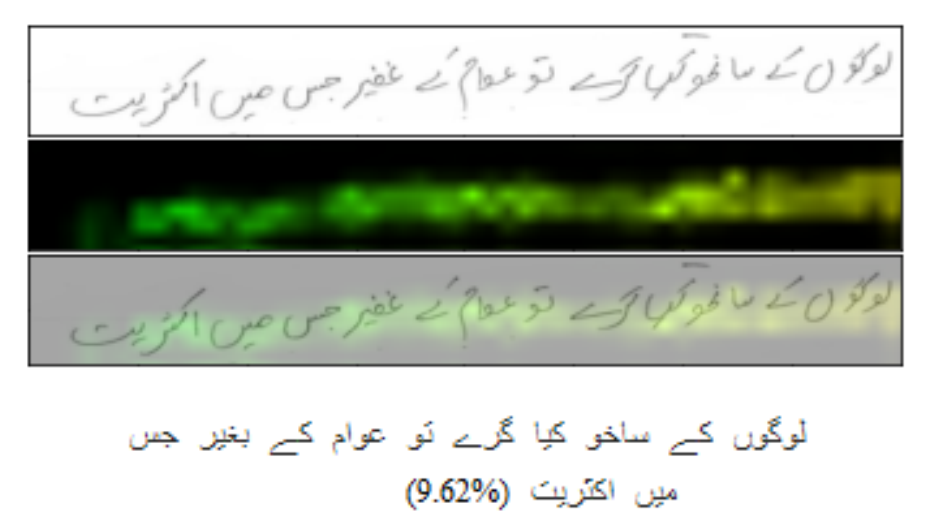}
        \\
        \includegraphics[width=.4\linewidth]{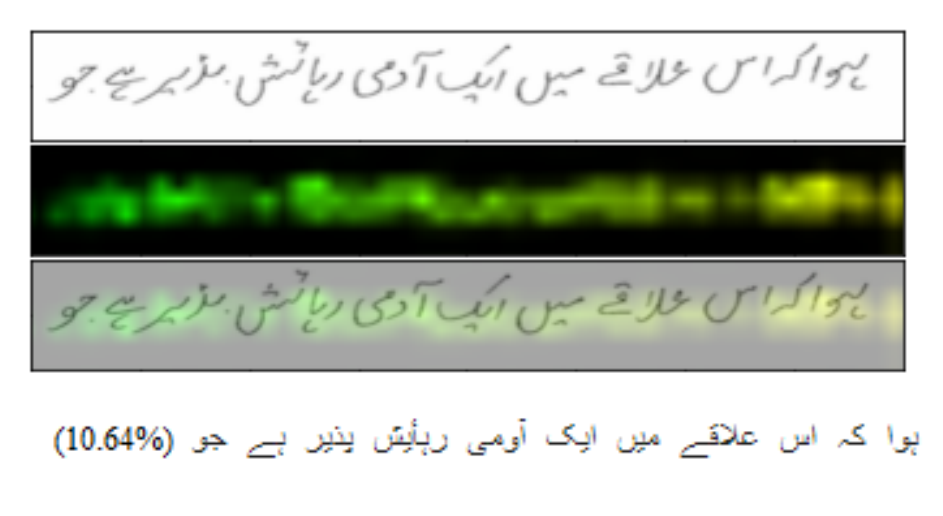} & \includegraphics[width=.4\linewidth]{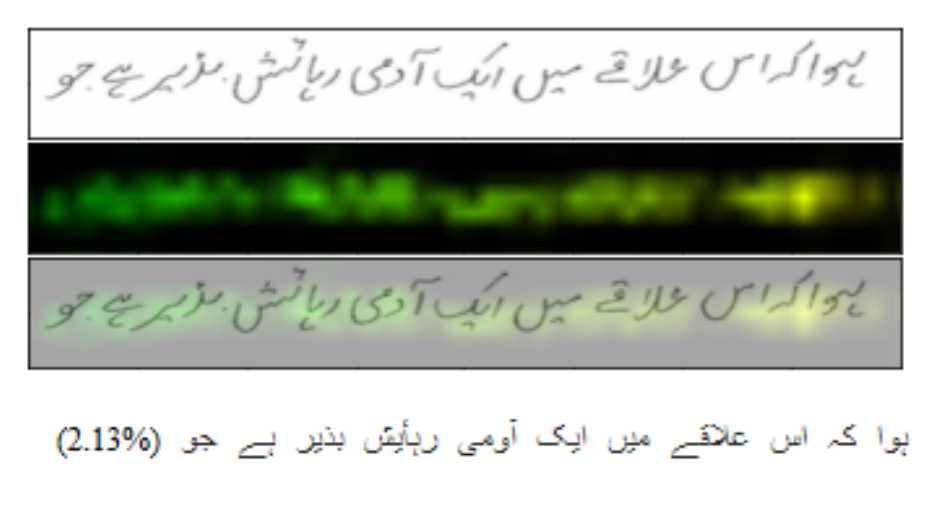}
        \\
        \includegraphics[width=.4\linewidth]{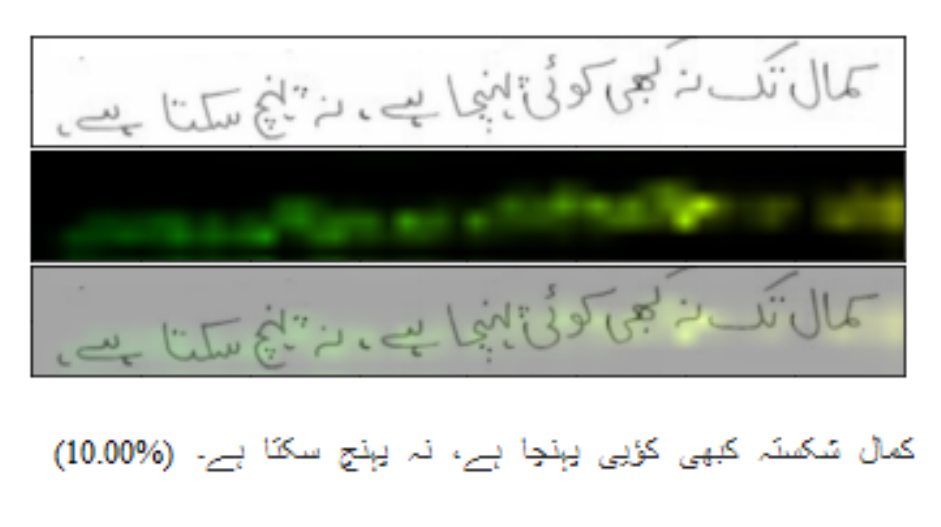} & \includegraphics[width=.4\linewidth]{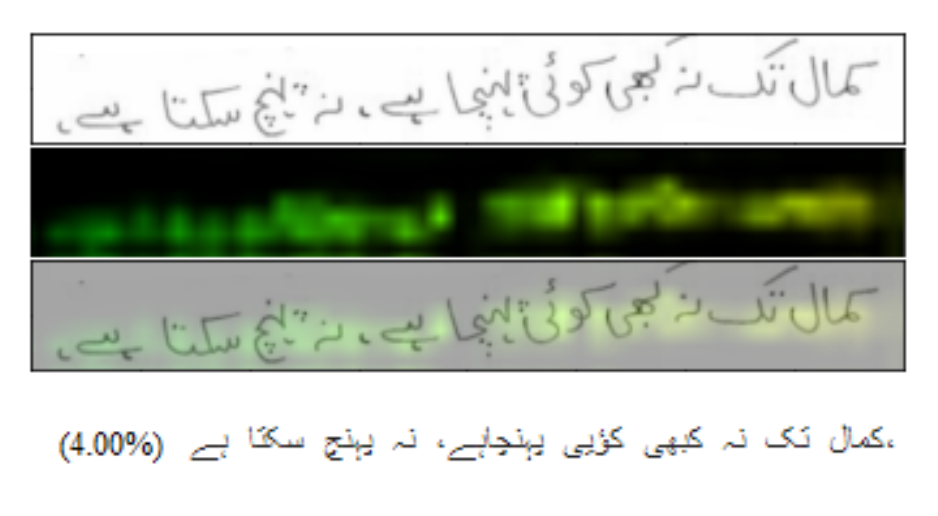}
    \end{tabular}
    \caption{Urdu recognition results. In each panel, top row is the input image, second row visualizes spatiotemporal attention, third row superimposes attention over the input image and fourth row is the recognized text string followed by the CER in brackets. The proposed CALText model learns to attend to localized regions of the image in context. The transition from yellow to green indicates a temporal sequence of spatial attention. The model learns to attend to Urdu text in a right-to-left direction. The model also \emph{follows} the text when the handwritten text is not exactly horizontal.}
    \label{fig:attention_results_urdu1}
\end{figure*}

\begin{figure*}
    \centering
    \begin{tabular}{cc}
        \textbf{Contextual Attention} & \textbf{Contextual Attention Localization}
        \\
        \includegraphics[width=.4\linewidth]{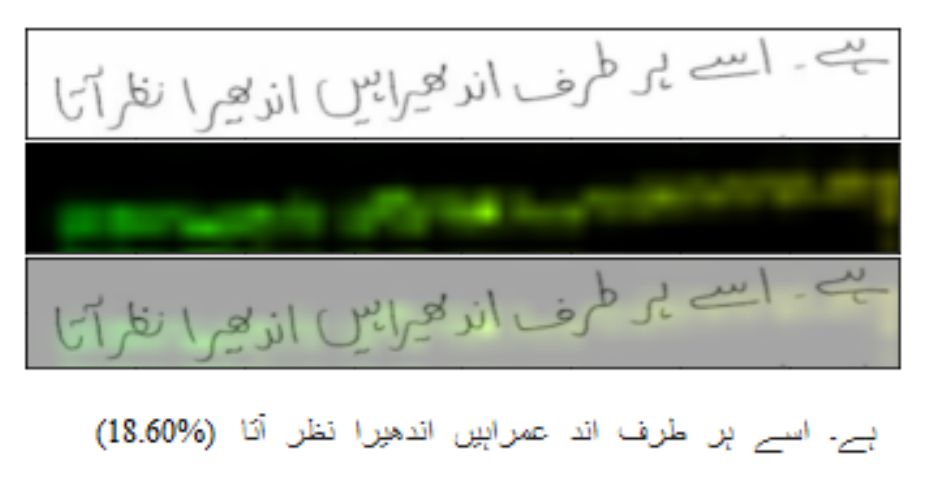} & \includegraphics[width=.4\linewidth]{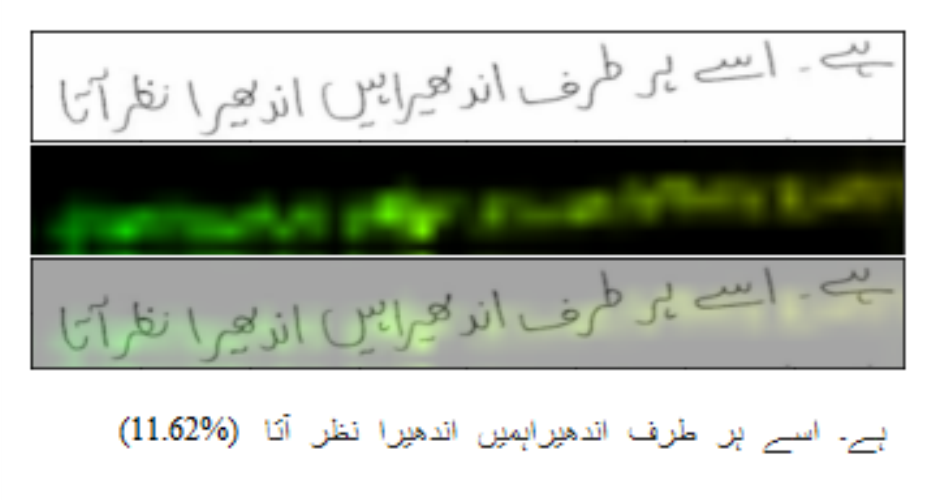}
        \\
        \includegraphics[width=.4\linewidth]{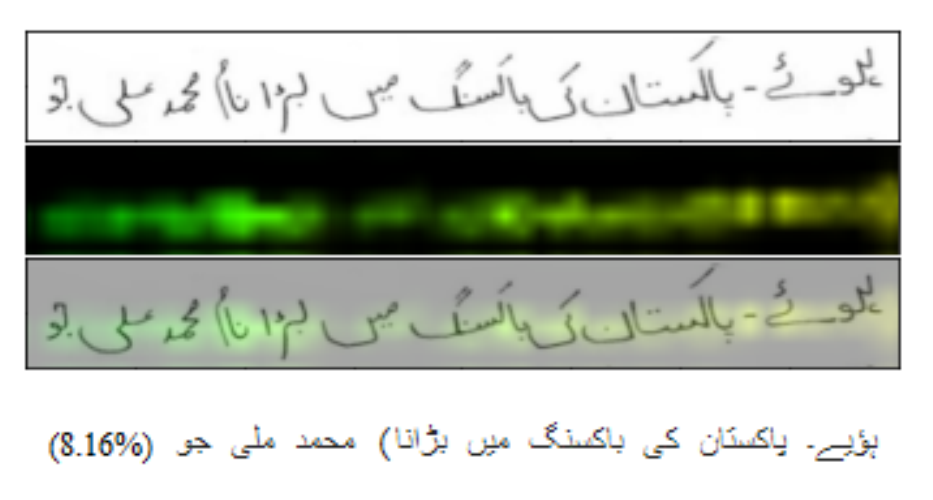} & \includegraphics[width=.4\linewidth]{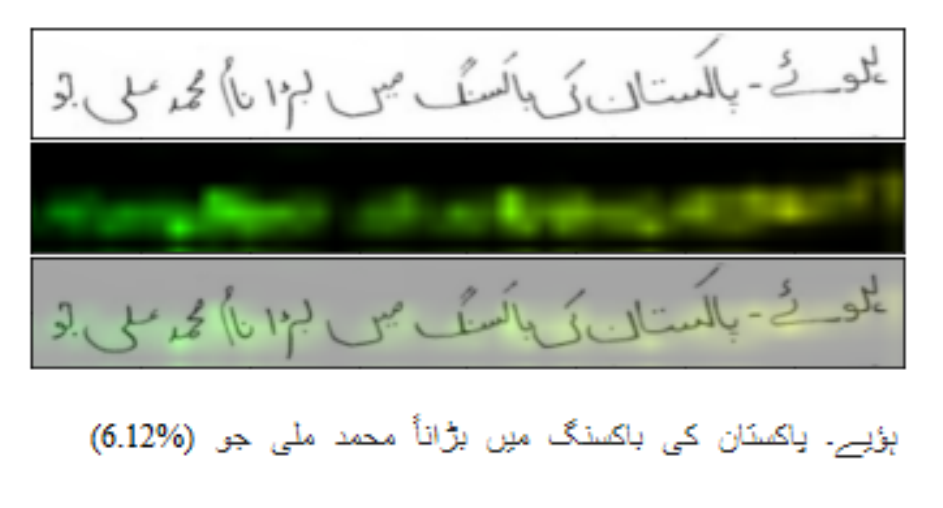}
        \\
        \includegraphics[width=.4\linewidth]{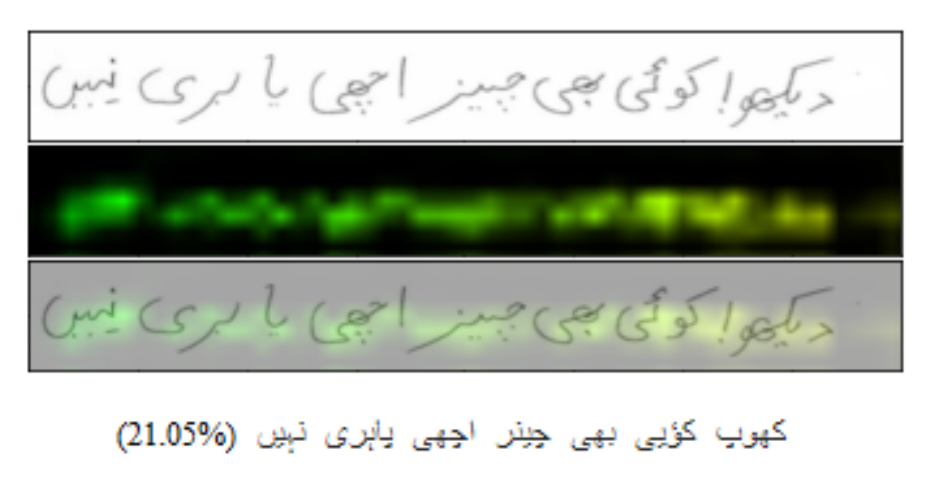} & \includegraphics[width=.4\linewidth]{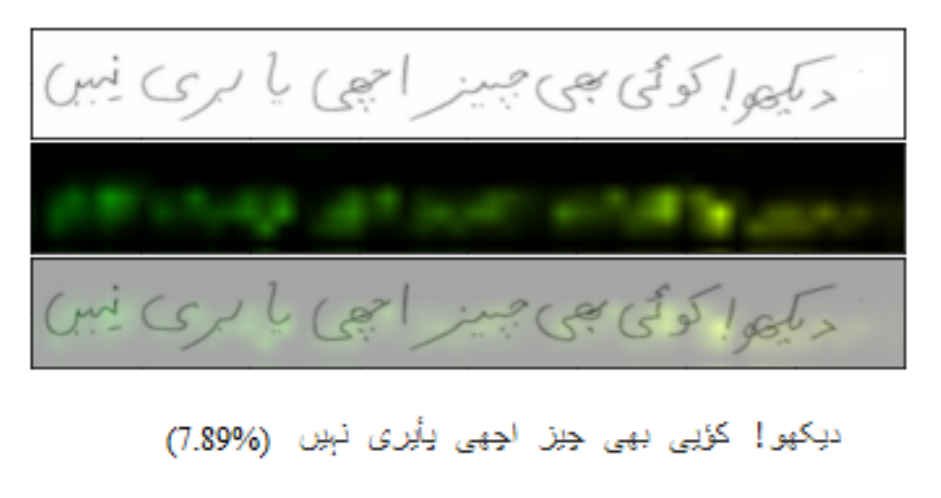}
        \\
        \includegraphics[width=.4\linewidth]{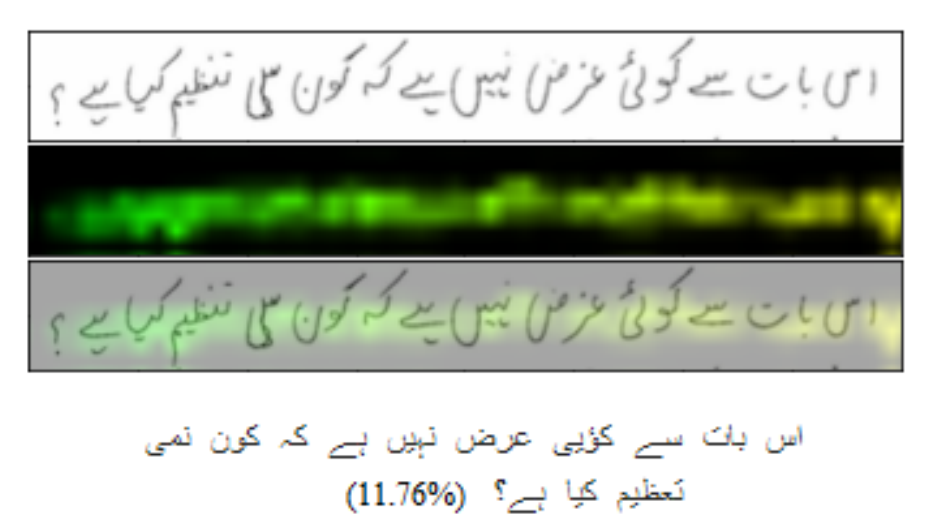} & \includegraphics[width=.4\linewidth]{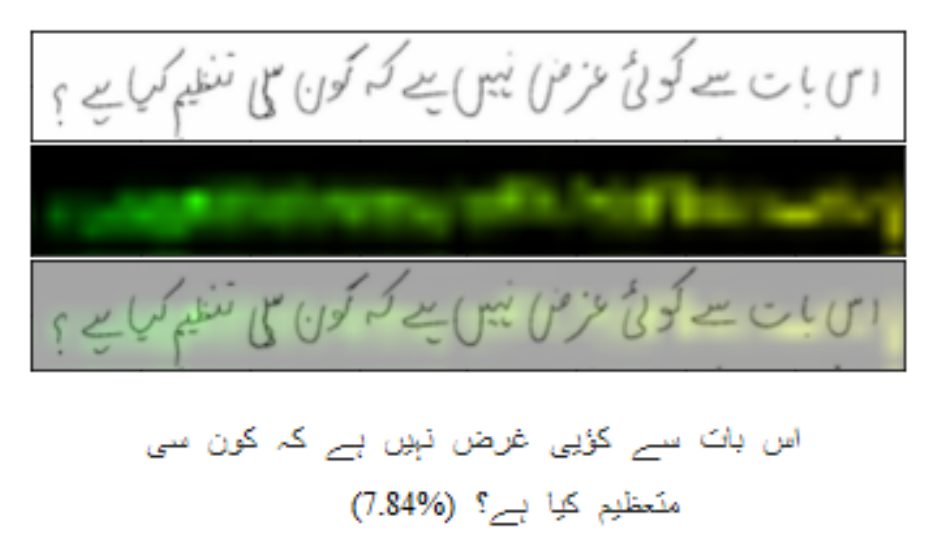}
        \\
        \includegraphics[width=.4\linewidth]{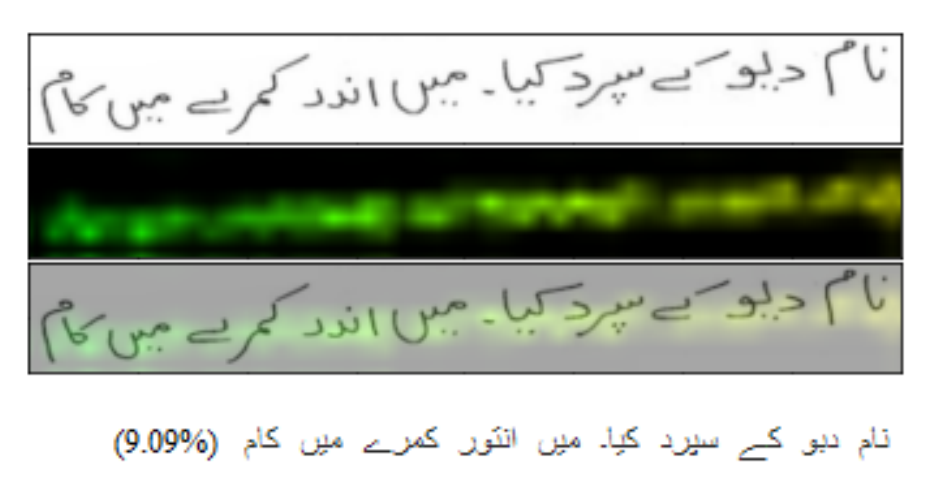} & \includegraphics[width=.4\linewidth]{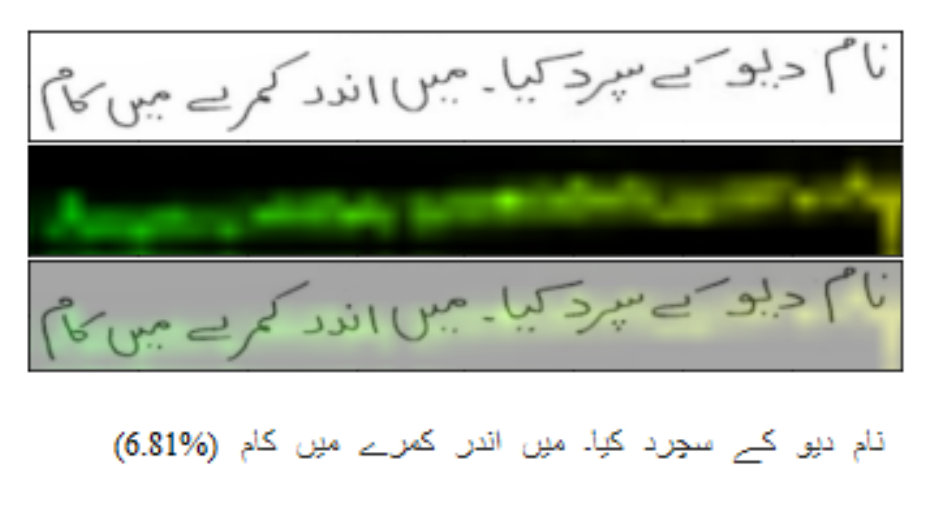}
    \end{tabular}
    \caption{Some more results on Urdu recognition.}
    \label{fig:attention_results_urdu2}
\end{figure*}

\begin{figure*}
    \centering
    \begin{tabular}{cc}
        \textbf{Contextual Attention} & \textbf{Contextual Attention Localization}
        \\
        \includegraphics[width=.4\linewidth]{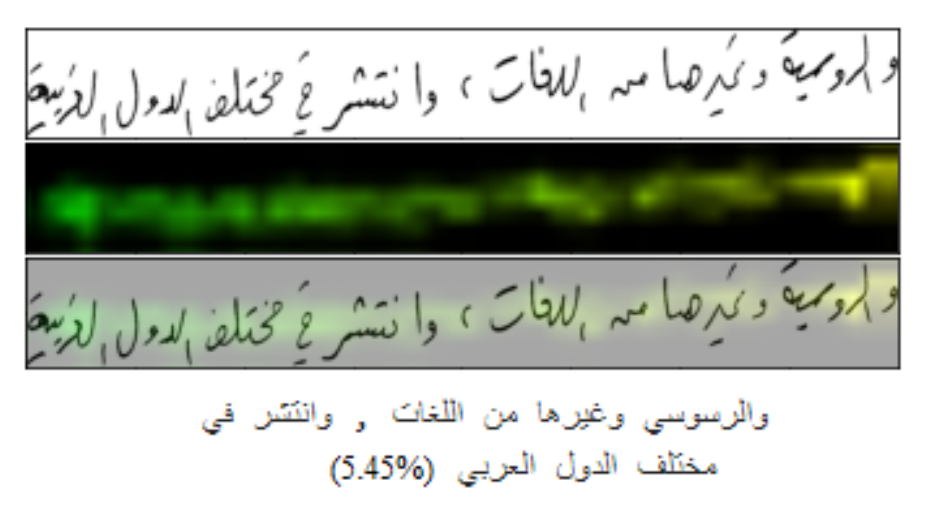} & \includegraphics[width=.4\linewidth]{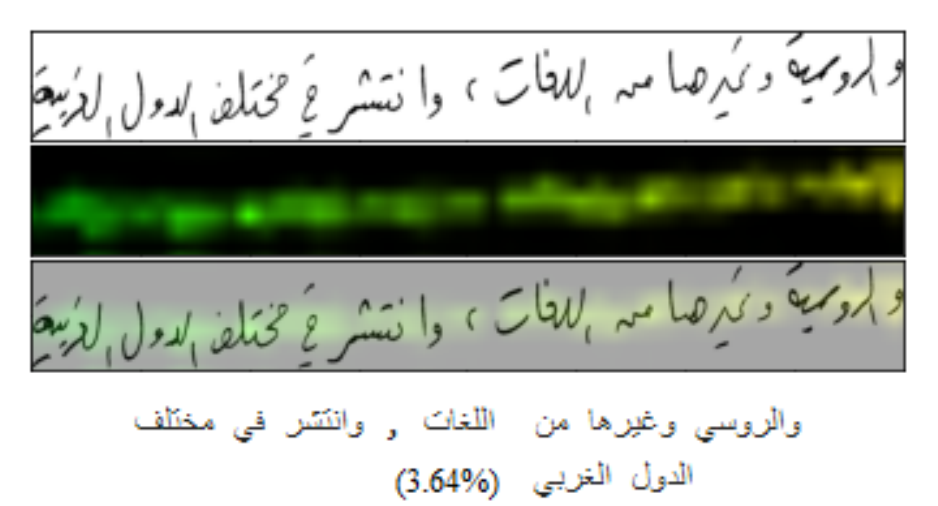}
        \\
        \includegraphics[width=.4\linewidth]{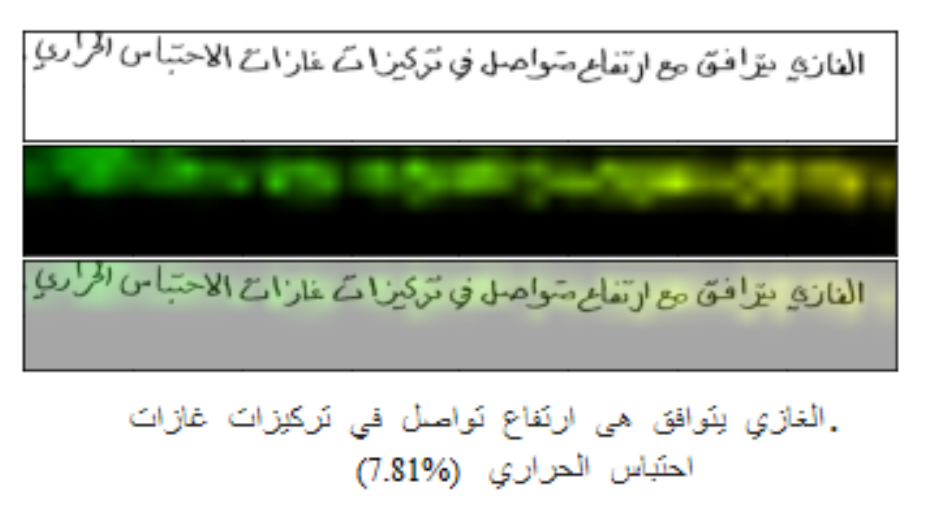} & \includegraphics[width=.4\linewidth]{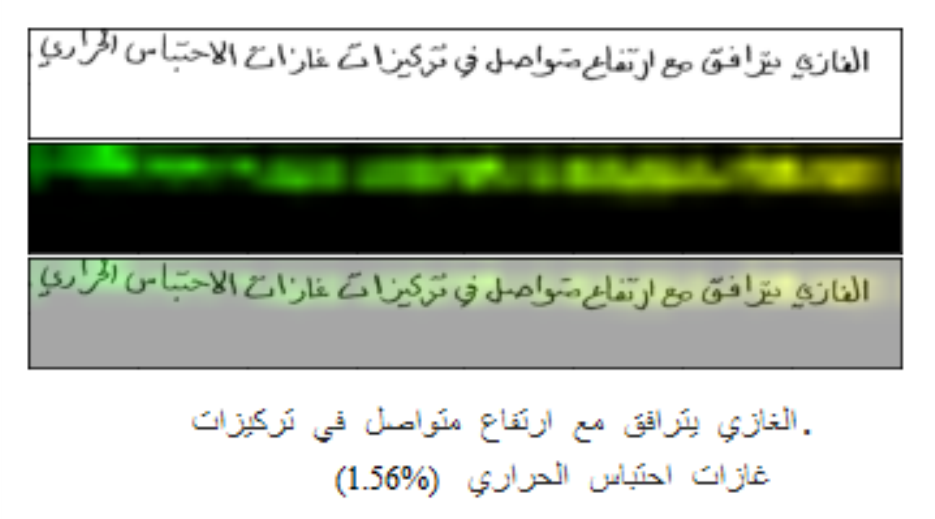}
        \\
        \includegraphics[width=.4\linewidth]{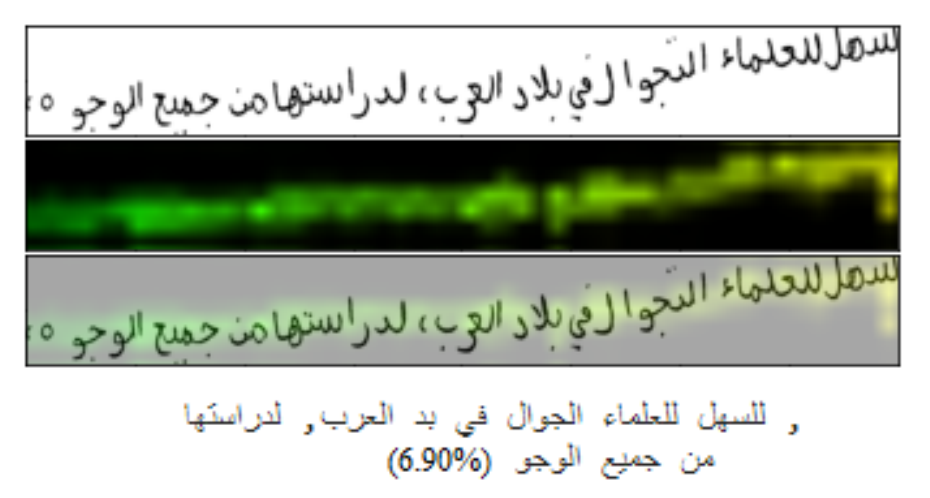} & \includegraphics[width=.4\linewidth]{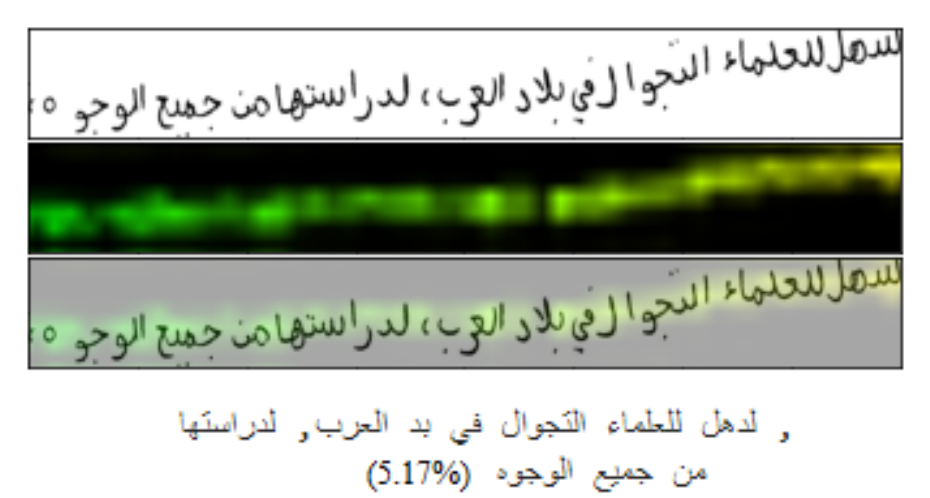}
        \\
        \includegraphics[width=.4\linewidth]{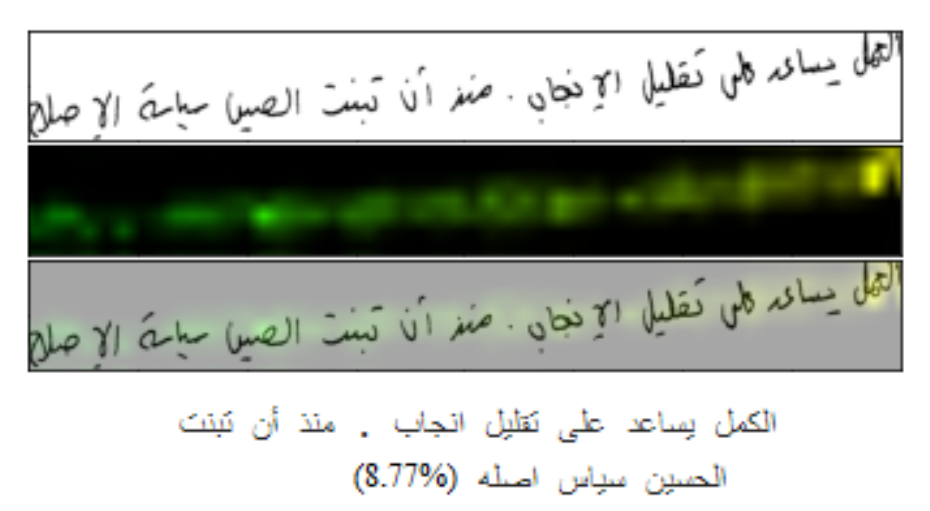} & \includegraphics[width=.4\linewidth]{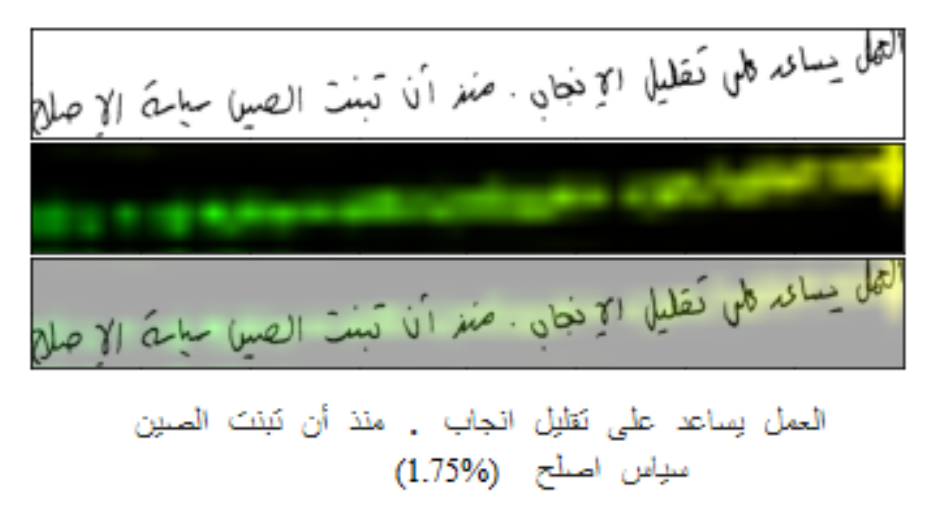}
    \end{tabular}
    \caption{The proposed CALText model learns to attend to localized regions of the image in context. The transition from yellow to green indicates a temporal sequence of attention. The model learns to attend to Arabic text in a right-to-left direction.
    }
    \label{fig:attention_results_arabic}
\end{figure*}

\begin{figure}[!t]
  \centering
  \captionsetup[subfigure]{labelformat=empty}
    \subfloat[][(a) Model trained for right-to-left recognition.]{
    \includegraphics[width=.45\linewidth]{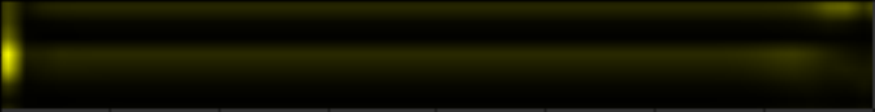}\label{fig:empty_r2l}
    }
    \subfloat[][(b) Model trained for left-to-right recognition.]{
    \includegraphics[width=.45\linewidth]{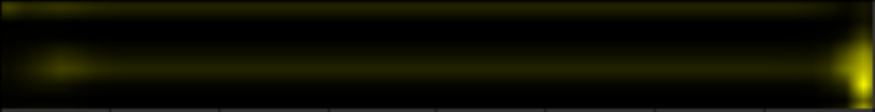}\label{fig:empty_l2r}
    }
\caption{Behavior of the proposed CALText model on an empty white image. Attention visualization for a model trained on (a) right-to-left text, and (b) left-to-right text. Both models output an empty string. As indicated by the yellow color, the outputs were obtained at the very first time step $t=1$. It can be noted that before outputting an empty string, the models focus more on the relevant end of the image. However, they do focus, albeit less, on the whole horizontal extant of the image as well. This is intuitively satisfying since the whole image \emph{should} be scanned before outputting an empty string.}
\label{fig:empty}
\end{figure}

\begin{figure}[!t]
    \centering
    \includegraphics[width=.45\linewidth]{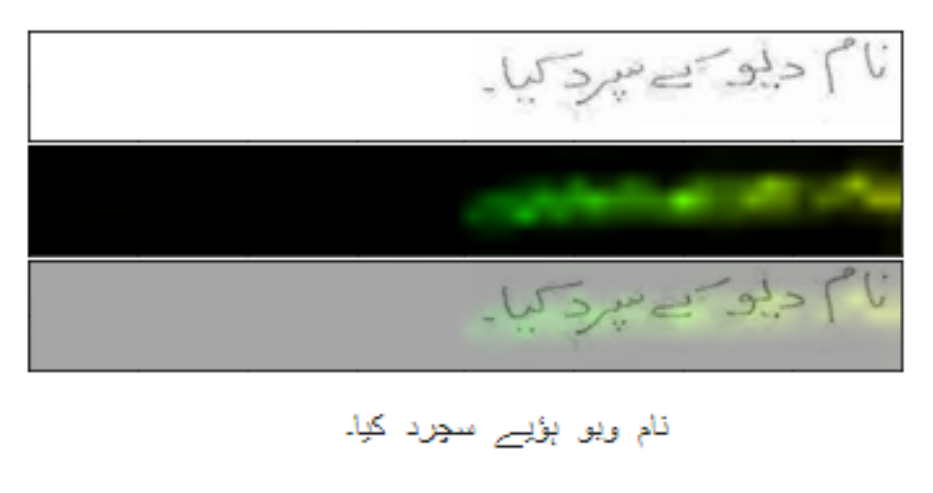}
    \includegraphics[width=.45\linewidth]{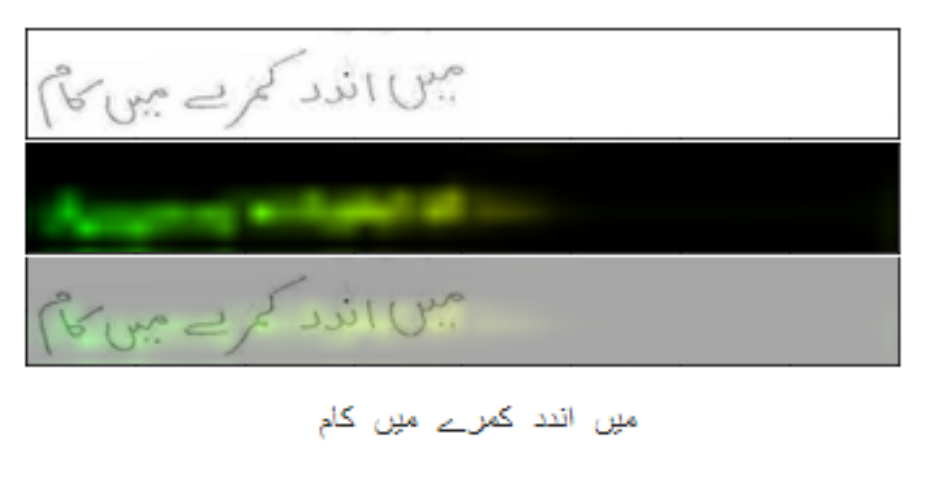}
    \caption{In order to produce its sequence of output characters, the proposed CALText model learns to attend to text regions exclusively.
    \label{fig:halfwhite}
    }
\end{figure}

A more compact visualization can be achieved by exploiting color to represent time. Let the optimal output after beam-search consist of $T$ steps. By dividing time by $T$ we can normalize it so that $0\leq t \leq1$. The normalized value of $t$ can be used to compute the color vector $\mathbf{c}_t$ at time $t$ by linearly interpolating between the color vector $\mathbf{c}_0$ at time $0$ and $\mathbf{c}_1$ at time $1$ as
\begin{align}
    \mathbf{c}_t &= (1-t)\mathbf{c}_0 + t\mathbf{c}_1
    \label{eq:color_interp}
\end{align}
The final colored attention through time can be visualized as
\begin{align}
    \tilde{\alpha} &= \sum_{t\in\{0,\dots,1\}} \mathbf{c}_t\alpha_t
    \label{eq:colored_spatiotemporal_attention}
\end{align}
The colored spatiotemporal attention image $\tilde{\alpha}$ visualizes attention both in terms of spatial intensity and temporal sequence. We set $\mathbf{c}_0$ equal to yellow and $\mathbf{c}_1$ equal to green. Therefore, in our visualizations, temporal sequence of attended locations moves from yellow in the beginning to green at the end. All subsequent visualizations (Figures \ref{fig:attention_results_urdu1} -- \ref{fig:halfwhite}) use this mechanism.

On handwritten Urdu text, Figures \ref{fig:attention_results_urdu1} and \ref{fig:attention_results_urdu2} demonstrate three properties of the proposed model.
\begin{enumerate}
    \item Encouraging the decoder to localize its attention reduces the character error rate.
    \item Temporally, learned attention moves from right-to-left which is the natural order of Urdu script.
    \item For skewed lines, attention follows the text.
\end{enumerate}
Figure \ref{fig:attention_results_arabic} demonstrates the same properties when our model is evaluated for handwritten Arabic text.

Figure \ref{fig:empty} demonstrates the regions attended by our model when observing an empty white image containing no text. The model trained for right-to-left ground-truth annotations correctly outputs an empty string as the recognition result at the first time step $t=1$ by focusing on the left-end of the image. In contrast, a model trained on left-to-right ground-truth annotations outputs an empty string after attending the right end of the image.

Figure \ref{fig:halfwhite} demonstrates that the proposed model has learned to attend directly to text. Instead of blindly scanning the image from left to right, it starts attending the beginning of the sentence text and proceeds until the end of the sentence text, irrespective of the spatial locations of the beginning and end.

\section{Conclusion}
\label{sec:conclusion}
We have proposed an encoder-decoder recognition of offline, handwritten text. The model has been evaluated for Urdu and Arabic which share numerous common features such as right-to-left ordering, similar characters and two-dimensional writing structure with inconsistent overlaps and spacing. Multiscale features are extracted through a DenseNet encoder. To incorporate the role of context in the process of \emph{reading} and to attend to specific characters, we have introduced a contextual attention localizer between two gated recurrent units. Results on test images show that our proposed CALText model learns to read and recognize text in a way that is similar to human reading by focusing only on relevant and localized image regions. The model determines relevancy of a region through contextual attention.

We have comprehensively re-annotated all the ground truth text lines of the PUCIT-OHUL dataset. As a result of our re-annotation, the count of unique characters increased from $98$ to $130$. Numerous mistakes and omissions in the original annotations have been corrected. For Urdu text, the use of contextual attention improves upon the state-of-the-art by a factor of $2\times$ in terms of both character and word level accuracy. Localization has been shown to further improve accuracies by around $2\%$. Code is made publicly available at \url{https://github.com/nazar-khan/CALText}.

\section*{Acknowledgements}
This research was supported by HEC-NRPU Grant 8029.

%%Harvard
\bibliographystyle{plain}
\bibliography{Bibliography}

\end{document}